\documentclass{article}

\PassOptionsToPackage{numbers, compress}{natbib}

    \usepackage[preprint]{neurips_2023}

\usepackage[utf8]{inputenc} 
\usepackage[T1]{fontenc}    
\usepackage{hyperref}       
\usepackage{url}            
\usepackage{booktabs}       
\usepackage{amsfonts}       
\usepackage{nicefrac}       
\usepackage{microtype}      
\usepackage{xcolor}         
\usepackage{CJKutf8}

\usepackage{hyperref}
\usepackage{url}

\usepackage[nolist]{acronym}

\makeatletter
\newcommand*{\org@overidelabel}{}
\let\org@overridelabel\@verridelabel
\@ifpackagelater{acronym}{2015/03/21}{
  \renewcommand*{\@verridelabel}[1]{%
    \@bsphack
    \protected@write\@auxout{}{\string\AC@undonewlabel{#1@cref}}%
    \org@overridelabel{#1}%
    \@esphack
  }%
}{
  \renewcommand*{\@verridelabel}[1]{%
    \@bsphack
    \protected@write\@auxout{}{\string\undonewlabel{#1@cref}}%
    \org@overridelabel{#1}%
    \@esphack
  }%
}
\makeatother

\usepackage{amsmath}
\usepackage{amsfonts}
\usepackage{amssymb}
\usepackage{wrapfig}
\usepackage{subcaption}
\usepackage{multirow}
 \usepackage{mathtools} 

\usepackage{verbatim}

\usepackage{anyfontsize}

\usepackage{microtype}
\usepackage{graphicx}
\usepackage{booktabs} 
\usepackage{multirow}
\usepackage{amsmath,amssymb}
\usepackage{booktabs}
\usepackage{caption,subcaption}

\usepackage{xcolor}
\definecolor{mygreen}{HTML}{3cb44b}
\definecolor{skyblue}{HTML}{beffff}
\definecolor{lightgreen}{HTML}{90ee90}

\usepackage{color, colortbl}

\definecolor{emerald}{rgb}{0.31, 0.78, 0.37}

\usepackage{tcolorbox}
\usepackage{enumitem}
\setitemize{itemsep=10pt,topsep=0pt,parsep=0pt,partopsep=0pt}
\usepackage{colortbl}

\usepackage{xcolor}
\definecolor{mygreen}{HTML}{3cb44b}
\colorlet{myyellow}{green!10!orange!90!}
\makeatletter

\usepackage{tikz}
\usetikzlibrary{arrows,shapes,snakes,automata,backgrounds,fit,petri}
\usepackage{adjustbox}

\newcommand{\RN}[1]{%
	\textup{\lowercase\expandafter{\it \romannumeral#1}}%
}
\usepackage{tabu}

\newcommand{\eg}[0]{\emph{e.g., }}

\newcommand{\beq}{\vspace{0mm}\begin{equation}}
\newcommand{\eeq}{\vspace{0mm}\end{equation}}
\newcommand{\beqs}{\vspace{0mm}\begin{eqnarray}}
\newcommand{\eeqs}{\vspace{0mm}\end{eqnarray}}
\newcommand{\barr}{\begin{array}}
\newcommand{\earr}{\end{array}}

\usepackage{color, colortbl}
\definecolor{Gray}{gray}{0.93}

\usepackage{lipsum}

\usepackage{pifont}
\newcommand{\cmark}{\ding{51}}%

\usepackage{makecell}

\newcommand*\circled[1]{\tikz[baseline=(char.base)]{
            \node[shape=circle,draw,inner sep=0.6pt] (char) {#1};}}

\usepackage{xcolor,amsmath}
\usepackage[linesnumbered,ruled,vlined]{algorithm2e}
\DontPrintSemicolon

\usepackage{xcolor}
\definecolor{mygreen}{HTML}{3cb44b}

\SetKwComment{Comment}{\color{green!50!black}\# }{}

\SetKwProg{Function}{def}{:}{}

\SetKwProg{For}{for}{:}{}
\SetKwProg{If}{if}{:}{}

\definecolor{GREEN}{RGB}{187, 255, 185}
\definecolor{RED}{RGB}{255,200,184}

\global
\tcbset{
  aibox/.style={
    width=\textwidth,
    top=10pt,
    colback=white,
    colframe=black,
    colbacktitle=black,
    enhanced,
    center,
    attach boxed title to top left={yshift=-0.1in,xshift=0.15in},
    boxed title style={boxrule=0pt,colframe=white,},
  }
}
\newtcolorbox{AIbox}[2][]{aibox,title=#2,#1}
\tcbset{
  aiboxsmall/.style={
    width=0.62\textwidth,
    top=10pt,
    colback=white,
    colframe=black,
    colbacktitle=black,
    enhanced,
    center,
    attach boxed title to top left={yshift=-0.1in,xshift=0.15in},
    boxed title style={boxrule=0pt,colframe=white,},
  }
}
\newtcolorbox{AIboxSmall}[2][]{aiboxsmall,title=#2,#1}
\definecolor{aigold}{RGB}{244,210, 1} 
\definecolor{aired}{RGB}{255,180,181}
\newlength\savewidth

\definecolor{defaultcolor}{gray}{0.9}

\newcolumntype{L}[1]{>{\raggedright\let\newline\\\arraybackslash\hspace{0pt}}m{#1}}
\newcolumntype{C}[1]{>{\centering\let\newline\\\arraybackslash\hspace{0pt}}m{#1}}
\newcolumntype{R}[1]{>{\raggedleft\let\newline\\\arraybackslash\hspace{0pt}}m{#1}}

\definecolor{paired-light-blue}{RGB}{198, 219, 239}
\definecolor{paired-dark-blue}{RGB}{49, 130, 188}
\definecolor{paired-light-orange}{RGB}{251, 208, 162}
\definecolor{paired-dark-orange}{RGB}{230, 85, 12}
\definecolor{paired-light-green}{RGB}{199, 233, 193}
\definecolor{paired-dark-green}{RGB}{49, 163, 83}
\definecolor{paired-light-purple}{RGB}{218, 218, 235}
\definecolor{paired-dark-purple}{RGB}{117, 107, 176}
\definecolor{paired-light-gray}{RGB}{217, 217, 217}
\definecolor{paired-dark-gray}{RGB}{99, 99, 99}
\definecolor{paired-light-pink}{RGB}{222, 158, 214}
\definecolor{paired-dark-pink}{RGB}{123, 65, 115}
\definecolor{paired-light-red}{RGB}{231, 150, 156}
\definecolor{paired-dark-red}{RGB}{131, 60, 56}
\definecolor{paired-light-yellow}{RGB}{231, 204, 149}
\definecolor{paired-dark-yellow}{RGB}{141, 109, 49}

\usepackage{fancybox} 
\usepackage{cleveref}
\usepackage{alltt}
\tcbuselibrary{most}
\usepackage{soul}

\usepackage{todonotes}

\usepackage{tikz}
\usetikzlibrary{shapes,arrows}

\usepackage{xcolor}

\newcommand{\shortname}{LLaVA-Interactive}

\title{\shortname{}: An All-in-One Demo for\\
Image Chat, Segmentation, Generation and Editing}

\author{%
  Wei-Ge Chen, Irina Spiridonova, Jianwei Yang, Jianfeng Gao, Chunyuan Li \\
  Microsoft Research, Redmond\\
  \texttt{\{wchen,irinasp,jianwyan,jfgao,chunyl\}@microsoft.com
  } \\  \\ 
\textcolor{magenta}{\url{https://llava-vl.github.io/llava-interactive/}}
  \\
}

\begin{document}

\maketitle

\begin{abstract}
  LLaVA-Interactive is a research prototype for multimodal human-AI interaction. The system can have multi-turn dialogues with human users by taking multimodal user inputs and generating multimodal responses. Importantly, \shortname{} goes beyond language prompt, where visual prompt is enabled to align human intents in the interaction. The development of LLaVA-Interactive is extremely cost-efficient as the system combines three multimodal skills of pre-built AI models without additional model training: visual chat of LLaVA~\cite{liu2023visual}, image segmentation from SEEM~\cite{zou2023segment}, and image generation and editing from GLIGEN~\cite{li2023gligen}. A diverse set of application scenarios is presented to demonstrate the promises of \shortname{} and to inspire future research in multimodal interactive systems.

\end{abstract}

\section{Introduction}

The rapid advancement of large language models (LLMs)~\citep{gpt4,gao2023llama} has revolutionized chatbot systems. As an example, OpenAI's ChatGPT~\citep{chatgpt} has demonstrated unprecedented levels of intelligence for human-AI interaction. The success of ChatGPT on language tasks has inspired the community to anticipate expanding the success to the multimodal space to eventually develop general-purpose multimodal AI agents~\citep{li2023multimodal}. 
The release of GPT-4V~\citep{gpt4v} is a big stride towards the goal. 
Although GPT-4V demonstrates many impressive AI skills~\cite{yang2023dawn,lu2023mathvista}, 
building multimodal conversational AI agents based solely on GPT-4V remains challenging for the open-source research community for two reasons. $(i)$ GPT-4V is primarily a language-based human-AI interaction system, where user-input images mainly provide visual contexts for text input and the system can only produce responses in text. $(ii)$ The details of model training and system architecture are not disclosed.

To mitigate the challenges, we present LLaVA-Interactive, an open-source research prototype system which can have multi-turn dialogues with human users by taking multimodal user inputs and generating multimodal responses. 
LLaVA-Interactive combines three multimodal skills of pre-built AI models without additional model training: visual chat of LLaVA~\cite{liu2023visual}, image segmentation from SEEM~\cite{zou2023generalized}, and image generation and editing from GLIGEN~\cite{li2023gligen}.
We hope that LLaVA-Interactive is complementary to GPT-4V for the development of future multimodal AI agents as LLaVA-Interactive provides a more extensible framework by supporting richer visual prompting and being open source. 

\begin{itemize}[leftmargin=7.5mm]
\setlength{\itemsep}{2pt}
\item 
\textit{Visual Prompting}. LLaVA-Interactive supports flexible language-vision human-AI interactions by allowing human users to use diverse visual prompts, such as drawing strokes, drag and drop or bounding boxes, to express user intents for completing sophisticated multimodal tasks that involve image segmentation, generation and editing. 
As a result, we find that compared to standalone LMMs, such as GPT-4V or LLaVA, LLaVA-Interactive can better follow user intents and generate more engaged human-machine interaction experiences.
\item 
\textit{Open-source}. We make our system and code base publicly available to facilitate future improvements in the community.
\end{itemize}

In the rest of this paper, Section 2 reviews related work. 
Section 3 describes the interface, workflow, and AI skills of \shortname{}. 
Section 4 presents a case study of developing an AI agent to assist photographic artists using \shortname{}. 
Section 5 presents a preliminary evaluation of \shortname{}.

\section{Related Works}

\paragraph{LMM with Visual Output and Interaction.} Most existing LMMs are developed to support visual chat -- image understanding and reasoning. There are several exploratory studies to enable LMM to support image output such as image generation/editing and segmentation, demonstrated in GILL~\citep{koh2023generating}, CM3leon~\citep{yu2023CM3Leon}, Emu~\citep{sun2023generative}, DreamLLM~\citep{dong2023dreamllm}, Kosmos-G~\citep{pan2023kosmos} and MGIE~\citep{fu2023guiding}. The idea is generalized to other modalities such as video and audio in NextGPT~\citep{wu2023next}. In contrast to model training to enable image output, another line of research is to prompt engineer LLM for multimodal tool use such as Visual ChatGPT~\cite{wu2023visual}, X-GPT~\cite{zou2023generalized}, MM-REACT~\cite{yang2023mm}, VisProg~\cite{gupta2022visual}, and ViperGPT~\cite{suris2023vipergpt}, where expert vision models with image output are activated in the inference time without any model training. Both research lines have demonstrated the extended capabilities with image output in LLMs. Similar to them, \shortname{} also supports image generation/editing and segmentation. \shortname{} is different from existing works in two aspects: 
$(i)$ \shortname{} is cheap in development, as it is a synergy of the inference stages of three models. There is no model training, and no prompt engineering of LLM. 
$(ii)$ Importantly, \shortname{} emphasizes the support of visual interaction, where a user can draw strokes to specify the human intent in segmentation and generation/editing, a unique capability that existing visual assistant systems do not have.

\paragraph{Unified Multimodal Modeling.} Inspired by the success of a single unified language model ChatGPT for language tasks, it is of great promise to build a general-purpose assistant with a single multimodal foundation model to complete more complex tasks~\citep{li2023multimodal}.
While the development of a unified LMM for all vision-language tasks is still exploratory~\citep{zou2023generalized,li2023uni,Gupta_2022_CVPR,chen2022unified,lu2022unified,wang2022ofa,wang2023seggpt}, it is believed that this direction has great promise in unlocking new application scenarios.
We present \shortname{} as a demo illustration of the potential of this research direction, including bridging vision and language tasks, completing a wide range of tasks in the wild, and providing a multimodal promptable user interface~\citep{li2023multimodal}.

\begin{figure}[t!]
\centering  
\vspace{-0mm}

  \centering  
  \begin{tabular}{ccc}  
    \multicolumn{3}{c}{\includegraphics[width=1.0\textwidth]{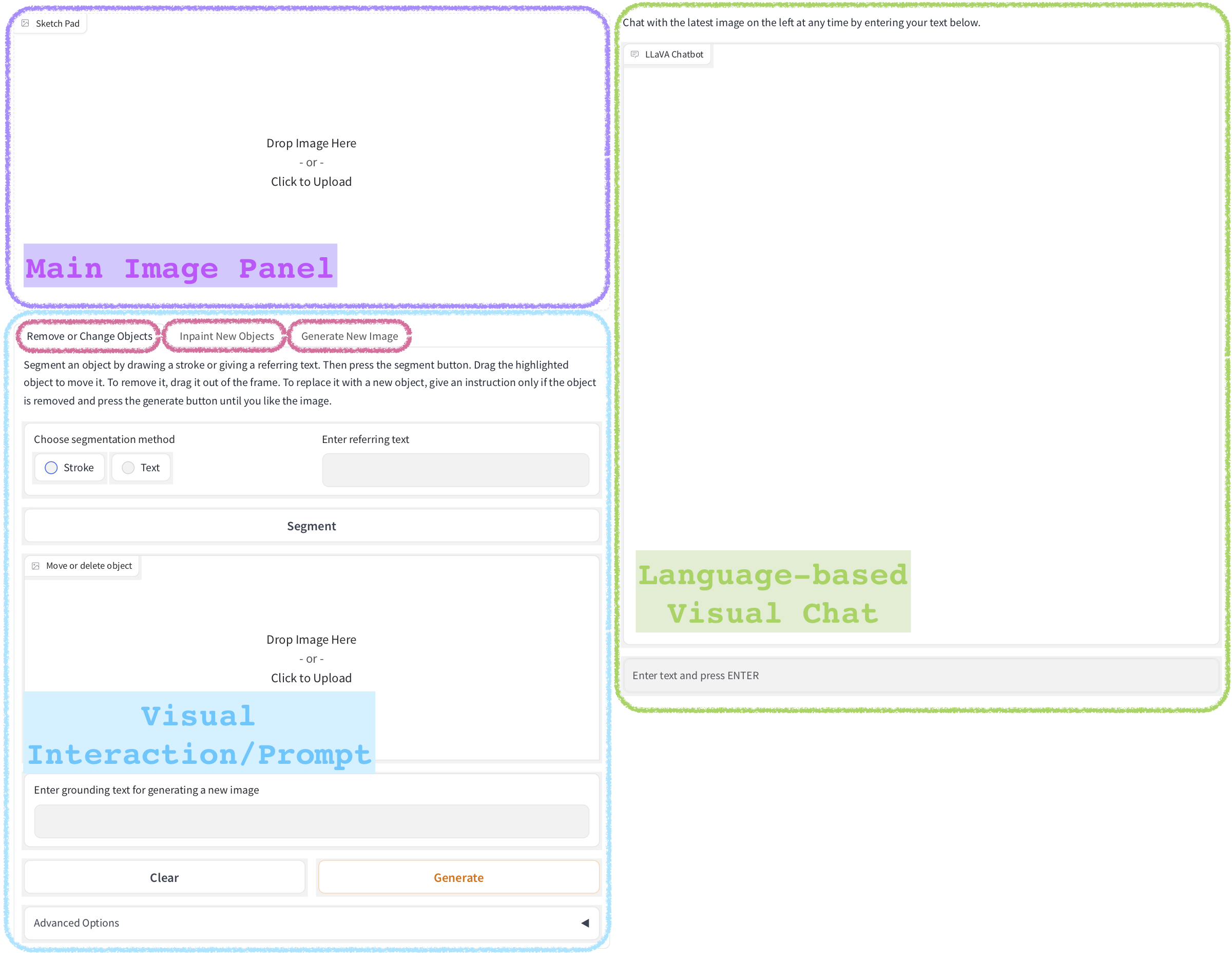}} \vspace{-1mm} \\ 
    \multicolumn{3}{c}{(a) The user interface layout. It consists of three panels, annotated in different colored for illustration.}  \vspace{2mm} \\ 
    \begin{tcolorbox}[colframe=black, colback=white, arc=4pt, width=0.31\textwidth, height=0.43\textwidth]  \includegraphics[width=1.1\textwidth]{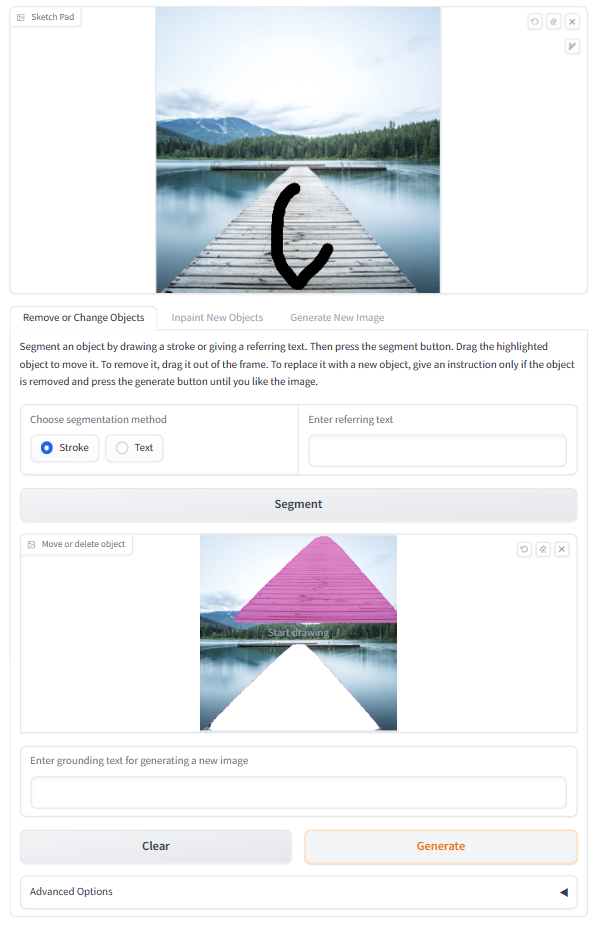}
    \end{tcolorbox}  &  
    \hspace{-1mm}
     \begin{tcolorbox}[colframe=black, colback=white, arc=4pt, width=0.31\textwidth, height=0.43\textwidth]
    \includegraphics[width=1.12\textwidth]{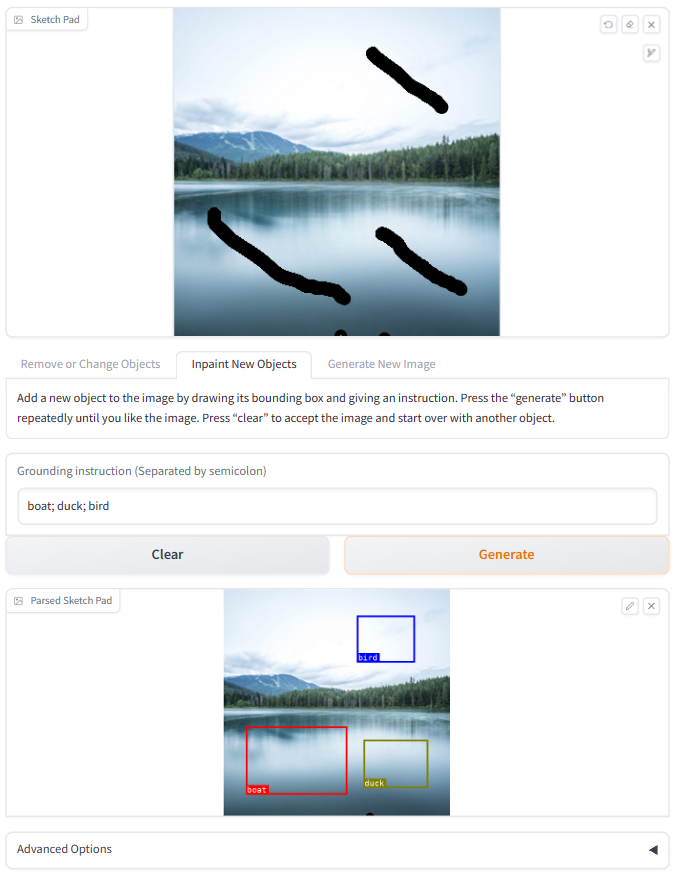} 
    \end{tcolorbox}  &  
    \hspace{-3mm}
     \begin{tcolorbox}[colframe=black, colback=white, arc=4pt, width=0.31\textwidth, height=0.43\textwidth]
    \includegraphics[width=1.12\textwidth]{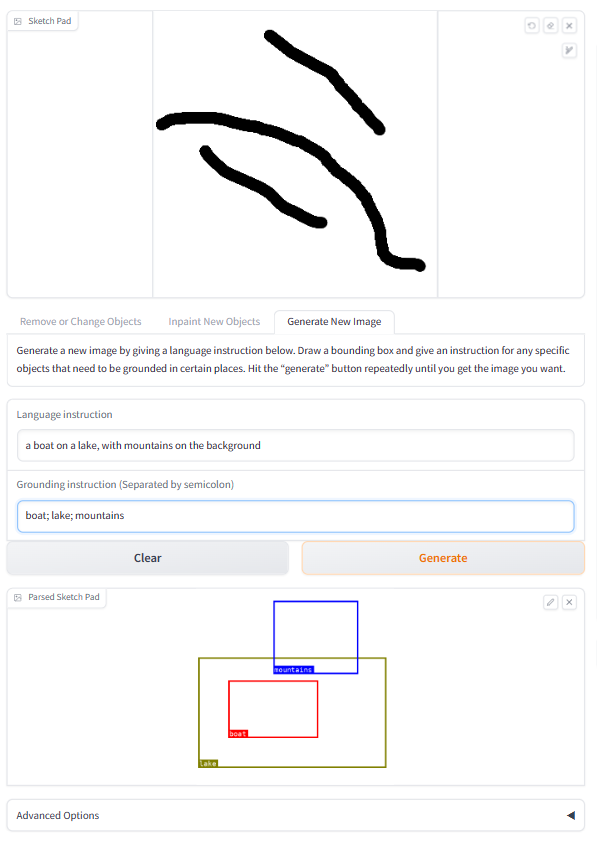}
    \end{tcolorbox}  \\ 
    \includegraphics[width=0.12\textwidth]{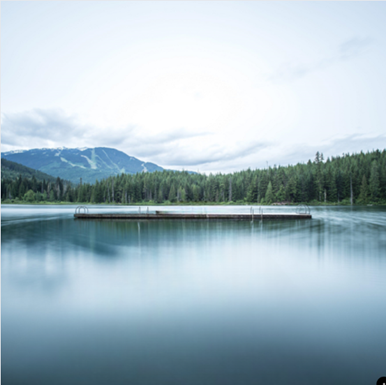} &  
    \includegraphics[width=0.12\textwidth]{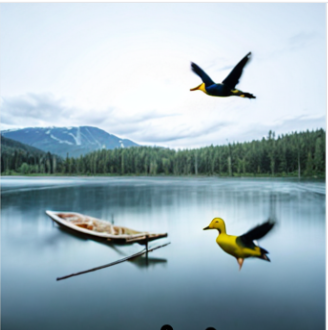} &  
    \includegraphics[width=0.12\textwidth]{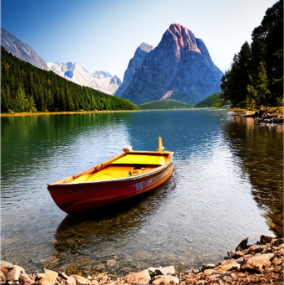} \\  
    (b) Remove or Change Objects &
    (c) Inpaint New Objects &
    (d) Generate New Image 
  \end{tabular}   
\vspace{-0mm}
\caption{The user interface of \shortname{}. (a) The overall user interface layout with three main panels, among which the visual interaction panel consists of three tabs based on their functionalities. They are shown in magenta rounded rectangles, and detailed in (b,c,d), whose image output after applying the visual interaction in each tab is shown at the bottom, respectively. }
\label{fig:llava_interactive_interface}
\end{figure}


\vspace{-0mm}
\section{LLaVA-Interactive}
\vspace{-0mm}
\paragraph{Interface.} Figure~\ref{fig:llava_interactive_interface} visualizes the user interface of \shortname{}. The overall interface layout is illustrated in (a), which consists of three panels annotated in different colors to ease the presentation. The top-left panel in purple maintains the up-to-date image and accepts visual prompts such as user strokes if necessary, the right panel in green is a language-based chat interface that accepts user questions about the image, and responds in natural language. The lower-left section, highlighted in blue, represents the visual interaction interface, which comprises three tabs. Each tab is distinguished by its unique function and is displayed within a red rounded rectangle.

To illustrate how humans can interact with \shortname{} using visual prompts, we provide one example for each tab in the sub-figures of Figure~\ref{fig:llava_interactive_interface}.
(b) {\it Remove or Change Objects.} For an image, the user draws a stroke on the object of interest. After clicking the ``\texttt{Segment}'' button, the object segmentation mask is provided, \eg the pier is highlighted in the magenta mask in this example. By dragging the mask out of the image, and clicking the ``\texttt{Generate}'' button, an edited image that removes the object is generated, \eg  the pier is removed in this example.
(c) {\it Inpaint New Objects.} To add objects with precise size and position in the image, the user could specify the object spatial configurations using bounding boxes. Each drawing corresponds a minimum size box that contains the stroke. The semantic concepts of the objects are provided in the grounding instruction (separated by a semicolon). By clicking the ``\texttt{Generate}'' button, the desired objects are inpainted in the input image, \eg a boat a duck are added in the lake, and a bird is added on the sky in this example.
(d) {\it Generate New Image.} To generate a new completely new image with precise object spatial layouts, one may specify the object layout using bounding boxes on ``\texttt{Sketch Pad}'', and provide the image-level caption as the language instruction.  By clicking ``\texttt{Generate}'' button, a new image that contains desired scene layout is generated. In this example, a new visual scene is generated to visualize the semantics of a boat on the lake, with mountains in the background.
At any given time, users can effortlessly switch among the three visual interaction tabs to fulfill their intended visual creation requirements iteratively.

\begin{figure}[t!]
\centering  
\vspace{-0mm}
\includegraphics[width=0.50\textwidth]{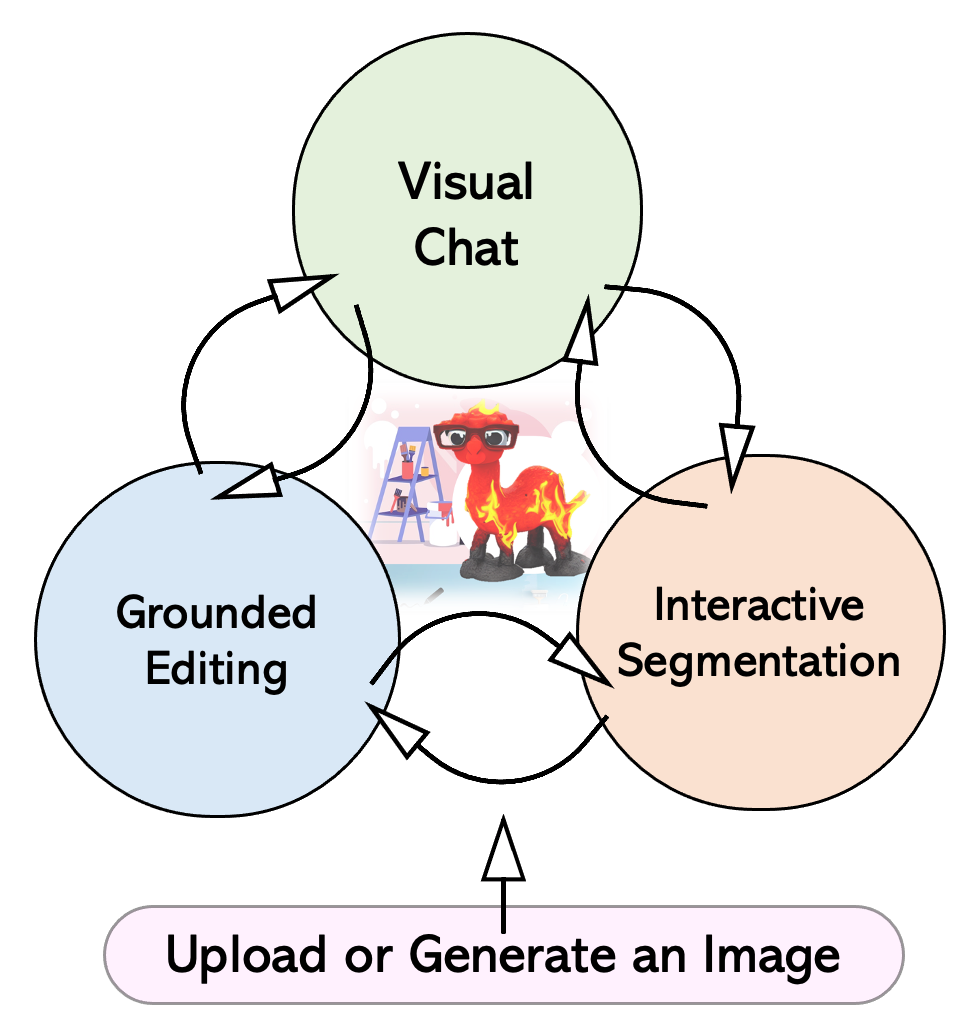} \\
\vspace{-0mm}
\caption{The workflow of \shortname{}.}
\label{fig:llava_interactive_workflow}
\end{figure}

\paragraph{Workflow.} 
Figure~\ref{fig:llava_interactive_workflow} provides a workflow of LLaVA-Interactive. We describe the typical visual creation process as below:

\begin{enumerate}[label=({\arabic*}),leftmargin=7.5mm]
\setlength{\itemsep}{2pt}
\item 
Image Input: To start, an image is required. The user can either upload an image or generate one by providing a language caption and drawing bounding boxes to establish the spatial arrangement of objects. Once the image is prepared, it can be interacted with through one of three methods: chat, segmentation, or editing.
\item 
Visual Chat: Users can ask questions about the image, such as seeking suggestions for revisions. Based on the editing recommendations, objects can be removed or added using Steps 3 or 4, respectively.
\item 
Interactive Segmentation: Users can create an object mask using either a stroke drawing or a text prompt. To remove the mask, drag it out of the image, and the background will automatically be filled. Alternatively, the mask can be moved to a different location. To replace the mask with a new object, provide a text prompt for the mask.
\item 
Grounded Editing: Users can directly place new objects on the image by drawing bounding boxes and associating the corresponding concepts with the intended objects.
\item 
Multi-turn Interaction: By repeating Steps 2, 3, or 4, users can iteratively refine their visual creations.
\end{enumerate}

\paragraph{Capability Comparisons}
Based on LLaVA that allows image input for visual chat only, LLaVA-Interactive extends the capabilities to support visual interaction such as user-drawn strokes and bounding boxes, as well as visual image generation/editing. Please see the comparisons of the capabilities below:

\begin{table}[h!]  
\centering  
\scalebox{1.0}{
\begin{tabular}{l|ccc}  
\toprule
System	& Visual Input	& Visual Output	& Visual Interaction \\
\midrule
LLaVA~\citep{liu2023visual} / GPT-4V~\citep{gpt4v}	& \cmark 		& &  \\
LLaVA-Interactive	& \cmark 	& \cmark 	& \cmark   \\
\bottomrule
\end{tabular}
}
\vspace{1mm}
\caption{Comparison with existing multimodal systems. The empty cells indicate inapplicable.
\vspace{-0mm}
}  
\label{tab:related_work_lmm}  
\end{table}

\subsection{Behind the Scenes: Individual Models}
\shortname{} is an all-in-one demo that connects three LV models in one interactive session for image chat, segmentation and generation/editing, which can complete more complex tasks than a single model alone. As a background, we briefly describe the individual models for those who are interested in the key techniques:
\begin{itemize}[leftmargin=7.5mm]
\setlength{\itemsep}{2pt}
\item 
\textbf{LLaVA}~\citep{liu2023visual}: Large Language and Vision Assistant, the first open-source alternative to GPT-4V. It is an end-to-end trained large multimodal model that combines CLIP vision encoder and Vicuna for general-purpose visual understanding and reasoning, achieving impressive chat capabilities mimicking the spirits of the GPT-4V. The recent LLaVA-1.5~\citep{liu2023improved} is considered in \shortname{}.
\item 
\textbf{SEEM}~\cite{zou2023segment}: Segment Everything Everywhere with Multi-modal prompts all at once. SEEM allows users to easily segment an image using prompts of different types including visual prompts (points, marks, boxes, scribbles) and language prompts. It can also work with any combination of prompts or generalize to custom prompts.
\item 
\textbf{GLIGEN}~\cite{li2023gligen}: Grounded-Language-to-Image Generation, an open-source model that extends the functionality of existing pre-trained text-to-image diffusion models by enabling them to also be conditioned on visual prompts such as bounding boxes.
\end{itemize}

\subsection{Development Challenges}
LLaVA-Interactive is a system-level demo synergy. It showcases the ability to create general-purpose assistants/agents by leveraging existing model checkpoints, eliminating the need for additional model training. While the training requirements for AI models are minimal, the development of LLaVA-Interactive presented various technical challenges that we addressed along the way. Firstly, we encountered difficulties with the GLIGEN inpainting model, which lacked the capability to fill background holes. As a solution, we incorporated the use of LaMA~\cite{suvorov2021resolution} for background filling. Secondly, Gradio, the framework used, lacked comprehensive support for user interaction, such as drag-and-drop functionality. To overcome this, we developed a new Gradio Image component tool that enabled the desired interaction. Additionally, integrating multiple complex projects and models posed a complexity challenge, which we addressed through experimentation, a streamlined UI layout, and an efficient data sharing scheme. Lastly, managing different package requirements and dependencies proved challenging, leading us to run separate web services for different models, such as LaMA.

\begin{figure}[h!]
\centering  
\vspace{-0mm}
\includegraphics[width=1.00\textwidth]{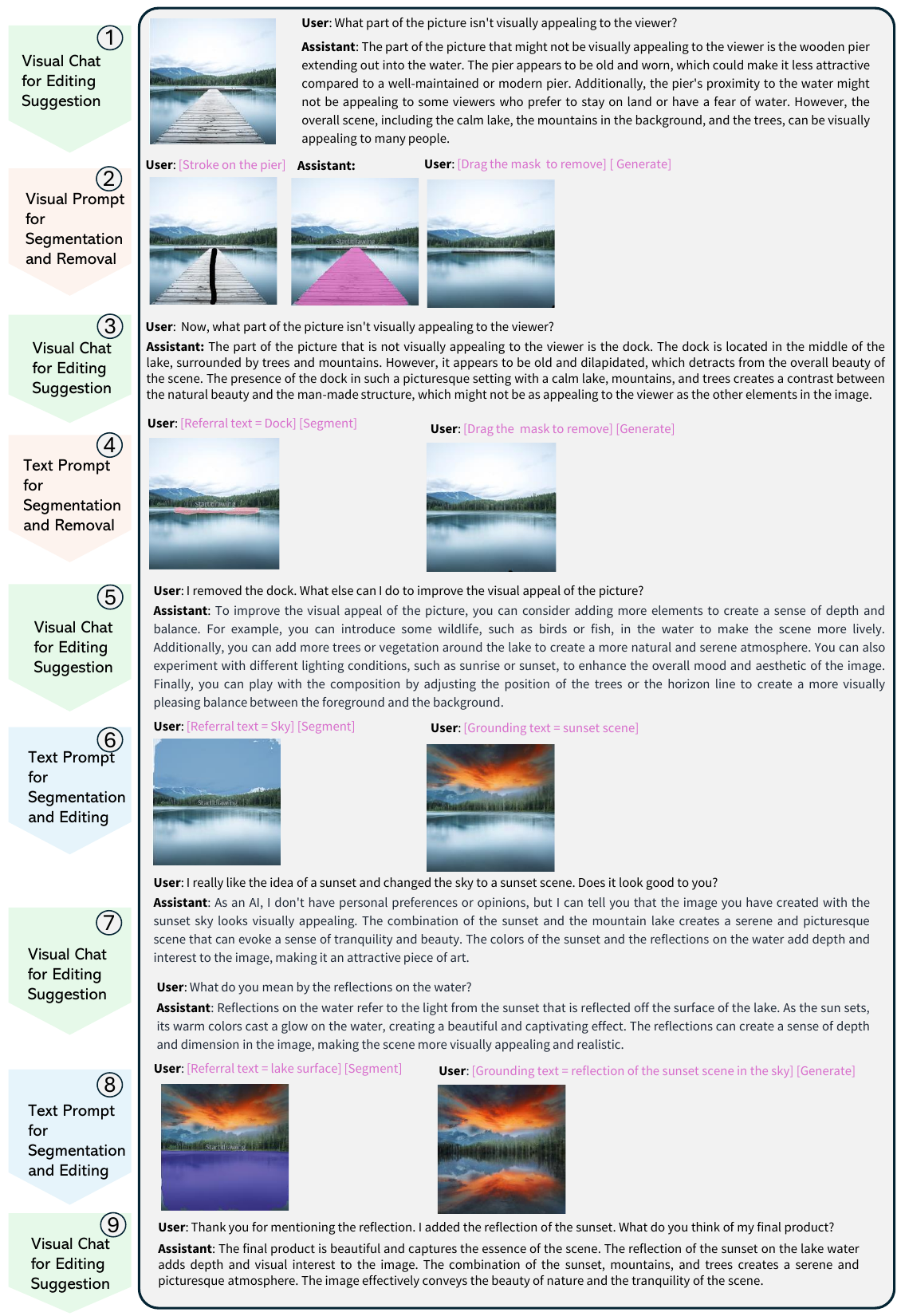} \\
\vspace{-0mm}
\caption{The use case study of \shortname{} as a photographic artist. The multi-turn multimodal interactive image editing process is illustrated. (Left) The interaction type/functionality for each turn; (Right) The detailed interaction process.}
\label{fig:lake_editing_example}
\end{figure}

\section{Case Study: Multimodal Interactive Creation for Photographic Artists}

To illustrate the enhanced user interaction experience and application scenarios made possible by \shortname{}, we present a case study focused on its utilization as a general-purpose assistant for photographic artists. Figure~\ref{fig:lake_editing_example} showcases a multi-turn multimodal interactive image editing process. The left side of the figure displays the interaction type and functionality for each turn, while the right side provides a detailed depiction of the interaction process. The language-based dialogue between the user and assistant is presented as text, with the user's visual prompts shown in magenta text and the resulting image editing outcomes displayed as processed images.

We describe the interaction process for this case study in Figure~\ref{fig:lake_editing_example}, which can be used as an example for the presentation of more application scenarios in Section~\ref{sec:application_scenarios}. 

\begin{enumerate}[label=\protect\circled{\arabic*},leftmargin=7.5mm]
\setlength{\itemsep}{2pt}
\item 
A user first uploads a lake scenery image, and asks for suggestions to improve the visual appearance. The assistant suggests to remove the wooden pier extending out into the water.
\item 
Following the suggestions, the user turns to the \ovalbox{\texttt{Remove and Change Object}} tab, chooses the \ovalbox{\texttt{Stroke}} mode, draws a stroke on the pier of the image, and clicks \ovalbox{\texttt{Segment}}. The segmentation mask of the pier is shown, presented in magenta color in this example. The user can further drag the mask out of the image, and click \texttt{Generate}, after which an updated image of the clear lake without the pier is shown. 
\item 
Based on the updated image, the user asks for further suggestions to improve the visual appearance. The assistant suggests to remove the small dock in the middle of the lake.
\item 
Following the suggestions, the user turns to the \ovalbox{\texttt{Remove and Change Object}} tab again. This time, the user utilizes the text prompt mode to segment the object, by choosing \ovalbox{\texttt{Text}}, typing the ``dock'' in the \ovalbox{\texttt{Enter referring text}} box, and clicking the \ovalbox{\texttt{Segment}} button.  The segmentation mask of the dock is shown, presented in magenta color. Following the same drag-and-generate procedure, the highlighted dock is removed from the image.
\item 
The user seeks further suggestions to improve the visual appeal of the picture, and the assistant recommends several concrete options, among which adding a sunset to enhance the overall aesthetic of the image is mentioned.
\item The user utilizes the text prompt mode again to select the sky region. To replace the selected mask with a new object, the user puts ``sunset scene'' in \ovalbox{\texttt{Enter grounding text for generating a new image}} box, and click \ovalbox{\texttt{Generate}}. A new image with a sunset in the sky is shown.
\item  The user asks for further editing suggestions, and finds that the reflection of the sunset on the water can make the image more appealing and realistic.
\item 
By performing a similar text prompt based segmentation and replacement procedure on the water, the reflection of the sunset scene in the sky is shown on the lake of the image.
\item The user asks for comments on the final product. The assistant considers this final version can effectively convey the beauty of nature.

\end{enumerate}

\section{Preliminary Evaluation with More Application Scenarios}
\label{sec:application_scenarios}

\subsection{Iterative Co-Creation of Visual Scene and Description}
\label{sec:image_generation_and_description}
\begin{figure}[!ht]
\begin{AIbox}{Iterative Visual Generation: Description and Editing (1/2)}

\scalebox{0.83}{
\begin{tabular}{p{1.3cm} p{13.2cm}  }
User & \textcolor{magenta}{(Generate New Image) [\texttt{Language instruction} = ``a boat on a lake, with mountains in the background''; Draw boxes for \texttt{Grounding instruction} ``boat; lake; mountains''] [Generate]}  \\
&  \includegraphics[height=6.0cm]{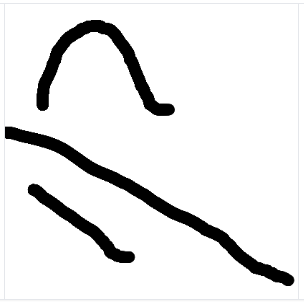} \\

Assistant & 
 \includegraphics[height=6.0cm]{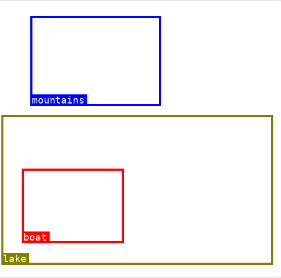} 
 \includegraphics[height=6.0cm]{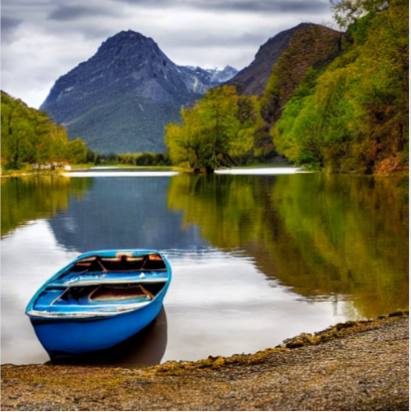}
  \\

User &  Please describe the image and promote the scenery. \\

Assistant & The image features a serene scene of a blue boat floating on a calm lake, surrounded by a beautiful mountain range. The boat is positioned near the shore, with the mountains in the background, creating a picturesque and tranquil atmosphere. The reflection of the mountains and the boat can be seen on the water's surface, enhancing the overall beauty of the scene. This idyllic setting is perfect for those seeking relaxation and a connection with nature. \\

User & \textcolor{magenta}{(Remove and Change Object) [Visual prompt by drawing a stroke on the shore][Generate]}\\
Assistant & 
 \includegraphics[height=4.0cm]{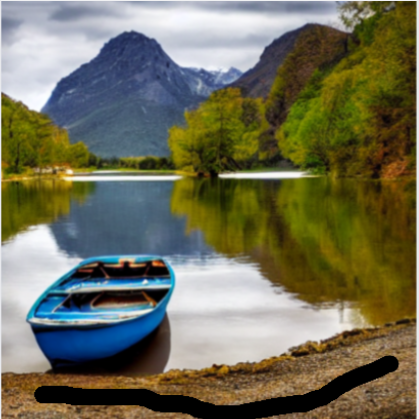} 
 \includegraphics[height=4.0cm]{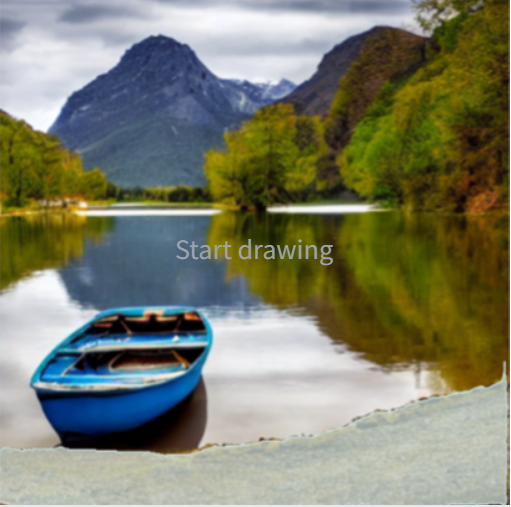} 
 \includegraphics[height=4.0cm]{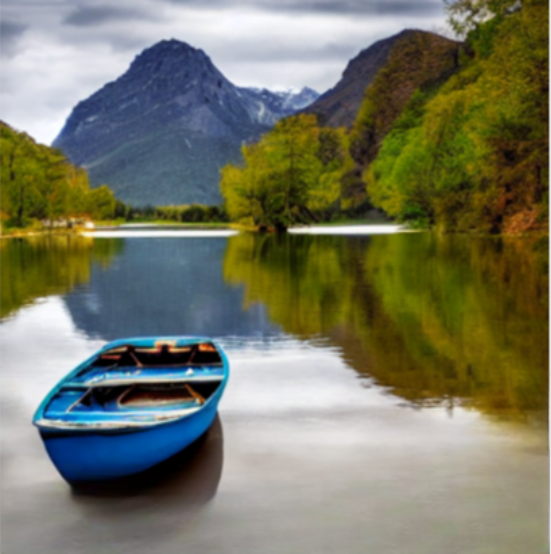} \\

User & \textcolor{magenta}{(Inpaint New Objects) [Draw boxes for \texttt{Grounding instruction} ``white goose''] [Generate]}  \\

Assistant & 
 \includegraphics[height=4.0cm]{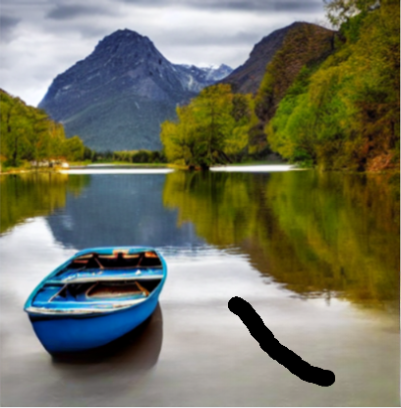} 
 \includegraphics[height=4.0cm]{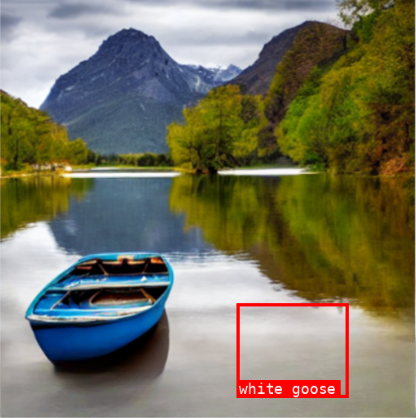} 
 \includegraphics[height=4.0cm]{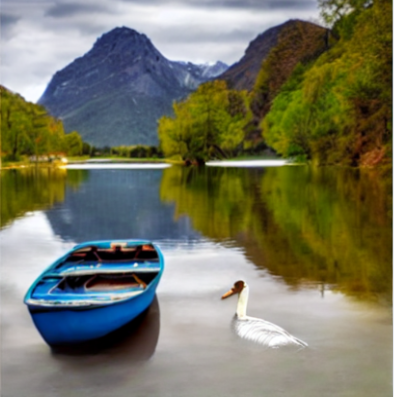}\\
\end{tabular}
}

\end{AIbox}
\vspace{-2mm}
\caption{
Iterative visual creation: image generation, description and editing (1/2). Refer to the scenario discussions in Section~\ref{sec:image_generation_and_description}.
} \label{fig:image_generation_and_description}
\vspace{-0mm}
\end{figure}

\begin{figure}[!ht]
\begin{AIbox}{Iterative Visual Generation: Description and Editing (2/2)}
\scalebox{0.83}{
\begin{tabular}{p{1.3cm} p{13.2cm}  }
 & 
 \includegraphics[height=6.0cm]{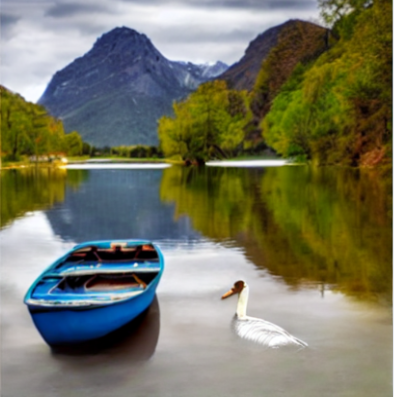} \\

User & write a poem about the image? \\
Assistant & 
A serene scene unfolds,
A blue boat rests on the water,
A swan swims by, gracefully,
In the calm lake, a peaceful sight.

The mountains rise in the distance,
A majestic backdrop to the scene,
The boat and swan coexist,
In this tranquil, picturesque setting.

The water's surface is still,
A reflection of the sky,
The boat and swan share the space,
In harmony, they glide.

The scene evokes a sense of calm,
A moment of respite from the world,
A peaceful coexistence,
Between the boat, swan, and the mountains. \\

User & 关于这一景色，写一首散文诗 \\
Assistant & 
在这一华丽的景色中，
一艘小船在湖面上漂浮，
一只优美的天鹅在水中游泳，
天空在水面上如画般反射，
天山在远方耸立，
宛如一幅丰收的画卷。

这一华丽的景色，
如诗如画，
湖泊的宁静，
船和天鹅的自由，
天山的壮丽，
都融入了这一美丽的画卷。

这一华丽的景色，
既有宁静，
也有自由，
也有壮丽，
它让我们感受到，
大自然的美妙与神秘。

这一华丽的景色，
是一首赞美大自然的诗，
是一幅描绘自然美景的画卷，
是一段感受自然的散文，
它都是一首赞美自然的赞歌。 \\

User & is the swan too big? \\
Assistant & 
In the image, the swan appears to be relatively large compared to the boat and the surrounding environment. However, it is not necessarily too big, as swans can vary in size depending on the species and individual. It is possible that the swan is simply larger than what one might expect in a typical lake or river setting. Nonetheless, the presence of a large swan in the image adds to the overall beauty and tranquility of the scene. \\

User & 
Please describe the image and promote the scenery \\
Assistant & 
The image features a serene and picturesque scene, with a boat floating on a body of water, such as a lake or river. A large swan is swimming nearby, adding a touch of grace and beauty to the landscape. In the distance, majestic mountains rise up, providing a stunning backdrop to the scene. The combination of the boat, swan, and mountains creates a sense of harmony and tranquility, inviting viewers to appreciate the natural beauty of the environment.

This idyllic setting is perfect for those seeking a peaceful escape from the hustle and bustle of everyday life. Whether it's for leisurely boating, birdwatching, or simply taking in the breathtaking views, this scenery offers a unique and memorable experience for anyone who visits. The image captures the essence of nature's beauty and the simple pleasures that can be found in the great outdoors.

\end{tabular}
}
\end{AIbox}
\vspace{-2mm}
\caption{
 Iterative visual creation: image generation, description and editing (2/2). Refer to the scenario discussions in Section~\ref{sec:image_generation_and_description}.
} \label{fig:image_generation_and_description_2}
\vspace{-0mm}
\end{figure}

Please see the user-assistant interaction in Figure~\ref{fig:image_generation_and_description} and~\ref{fig:image_generation_and_description_2}. Joint creation of a visual scene and its associated text from scratch can be useful for content creators. In this scenario, the user can generate a peaceful and relaxing outdoor scene from text, and specify spatial layout by drawing boxes. Once the image is generated, the users can ask the assistant to describe and promote the image using text. The visual image creation in one shot can be imperfect, and can be iteratively improved, \eg removing the gray shore and adding a white goose using a visual prompt in this example. For the final image, the user can ask the assistant to produce a poem in both English and Chinese to promote the image. If necessary, the user can also ask if the synthsized image is reasonable or not, \eg the size of the swan compared with the boat. It can be seen the text description and response of \shortname{} is often coherent with edited images.

\subsection{Graphic Design for a Halloween Poster}
\label{sec:graphics_design_halloween}
\begin{figure}[!ht]
\begin{AIbox}{Graphic Design for a Halloween Poster. (1/3) }
\scalebox{0.83}{
\begin{tabular}{p{1.3cm} p{13.2cm}  }

User &  I'd like to design a poster for Halloween. Can you give me some ideas? \\
Assistant & 
    Sure! Here are a few ideas for a Halloween poster:
    \begin{itemize}
            \item A spooky scene with a full moon, a graveyard, and some ghosts or other supernatural creatures.
            \item A pumpkin patch with carved pumpkins and a scarecrow.
            \item A haunted house with a creepy clown or other scary characters lurking inside.
            \item A group of trick-or-treaters out on Halloween night, with a mix of cute and spooky elements.
            \item A witch's cauldron with ingredients for a potion, such as bats, spiders, and ghostly eyeballs.
            \item A haunted forest with glowing eyes and twisted trees.
            \item A spooky lab with a mad scientist and his monstrous creations.
            \item A graveyard at night with glowing headstones and a ghostly figure.
            \item A haunted mansion with a ghostly family and their spooky guests.
            \item A group of skeletons having a party, with skulls and bones as decorations.
    \end{itemize}            
    I hope these ideas give you some inspiration for your Halloween poster! \\

User & \textcolor{magenta}{(Generate New Image) [\texttt{Language Instruction } = "A pumpkin patch with carved pumpkins and a scarecrow"] [Generate]} \\
Assistant &   \includegraphics[height=6.0cm]{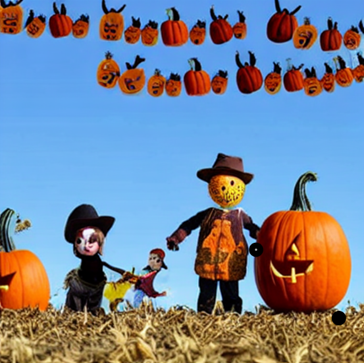} \\

User &  What do you think of this picture I generated? Do you have any ideas to improve? \\
Assistant & 
    The image features a group of scarecrows dressed in Halloween costumes, standing in a field with pumpkins. There are several pumpkins scattered throughout the scene, with some placed closer to the scarecrows and others further away. The scarecrows are positioned in various poses, creating a lively and festive atmosphere.
    The scene is set against a blue sky, which adds a sense of depth and contrast to the image. The combination of scarecrows, pumpkins, and the blue sky creates a visually appealing and spooky Halloween scene.
    To improve the image, one could consider adding more elements to the scene, such as a few more pumpkins in different sizes and positions, or incorporating some additional Halloween-themed decorations like bats, spiders, or ghostly figures. This would create a more dynamic and engaging visual experience for the viewer. \\

\end{tabular}
}
\end{AIbox}
\vspace{-2mm}
\caption{
 Graphic Design for a Halloween Poster (1/3). Refer to the scenario discussions in Section~\ref{sec:graphics_design_halloween}.
} \label{fig:graphics_design_halloween_1}
\vspace{-0mm}
\end{figure}

\begin{figure}[!ht]
\begin{AIbox}{Graphic Design for a Halloween Poster. (2/3) }
\scalebox{0.83}{
\begin{tabular}{p{1.3cm} p{13.2cm}  }

User & \textcolor{magenta}{(Inpaint New Objects)[Draw box for \texttt{Grounding Instruction} = "bat"][Generate]} \\
Assistant & \includegraphics[height=6.0cm]{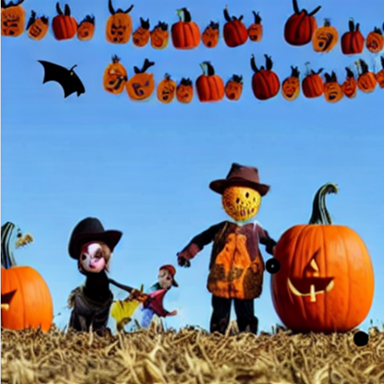} \\

User & What do you think of the bat I added? Do I need more of them for this Halloween poster? \\
Assistant &
    The bat you added to the Halloween poster is a nice touch, but it might not be enough to create a more spooky atmosphere. To enhance the poster's Halloween theme, you could consider adding more bats, spiders, or other scary elements such as ghosts, skeletons, or cobwebs. This would create a more visually engaging and thrilling poster that captures the essence of Halloween. \\

User & \textcolor{magenta}{(Change or Remove Objects) [Visual prompt by stroking  on one scarecrow][Segment][\texttt{Grounding Instruction} = "Halloween ghost"][Generate]} \\
Assistant &  \includegraphics[height=6.0cm]{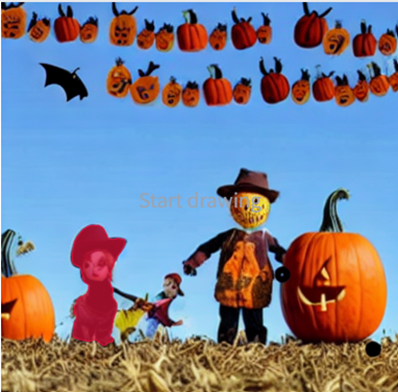} \includegraphics[height=6.0cm]{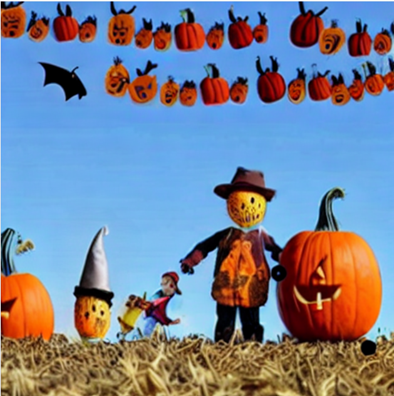} \\

\end{tabular}
}
\end{AIbox}
\vspace{-2mm}
\caption{
 Graphic Design for a Halloween Poster (2/3). Refer to the scenario discussions in Section~\ref{sec:graphics_design_halloween}.
} \label{fig:graphics_design_halloween_2}
\vspace{-0mm}
\end{figure}

\begin{figure}[!ht]
\begin{AIbox}{Graphic Design for a Halloween Poster. (3/3) }
\scalebox{0.83}{
\begin{tabular}{p{1.3cm} p{13.2cm}  }

User & \textcolor{magenta}{(Remove or Change Objects)[Visual prompt by stroking on the small scarecrow][Segment][Visual prompt by dragging out of picture to remove] [Generate]} \\
Assistant & \includegraphics[height=4.0cm]{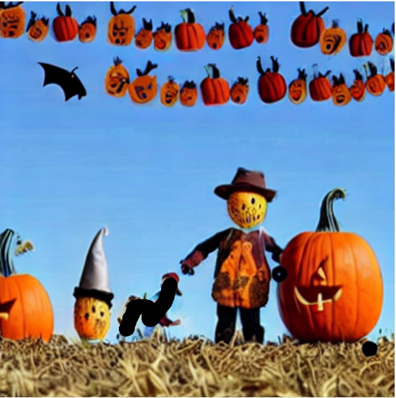} \includegraphics[height=4.00cm]{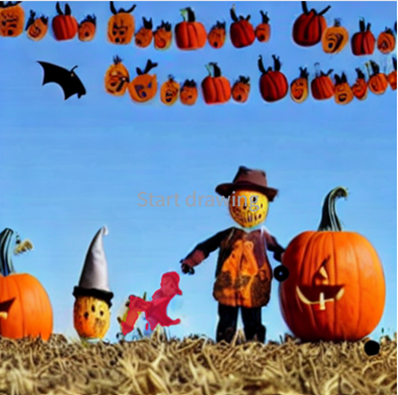} 
\includegraphics[height=4.0cm]{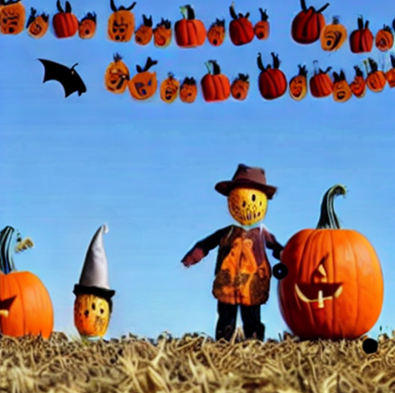}  \\

User & \textcolor{magenta}{(Inpaint Objects) [Draw box for \texttt{Grounding Instruction} = ``Halloween skeleton''][Generate]} \\
Assistant & \includegraphics[height=4.0cm]{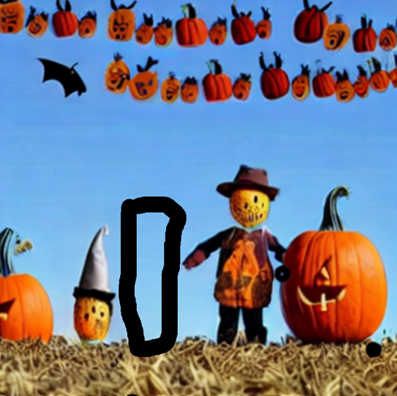} \includegraphics[height=4.0cm]{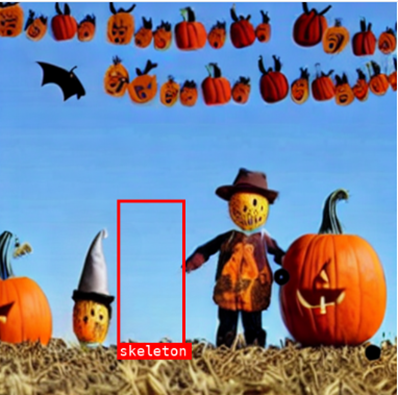} 
\includegraphics[height=4.0cm]{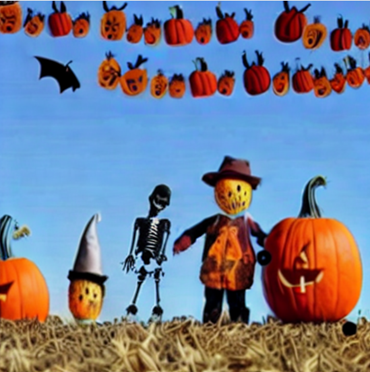}  \\

User & \textcolor{magenta}{(Inpaint Objects) [Clear to accept][Draw box for \texttt{Grounding Instruction} = "Halloween spider web"][Generate]} \\
Assistant & \includegraphics[height=4.0cm]{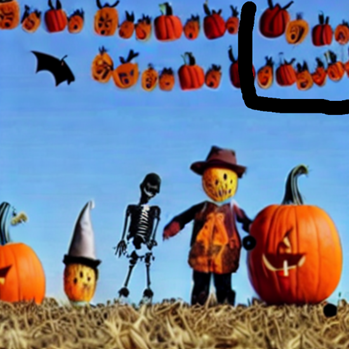} \includegraphics[height=4.0cm]{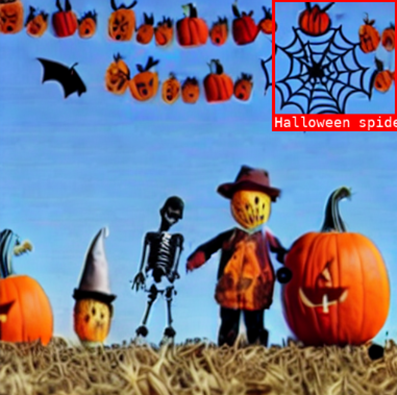} 
\includegraphics[height=4.0cm]{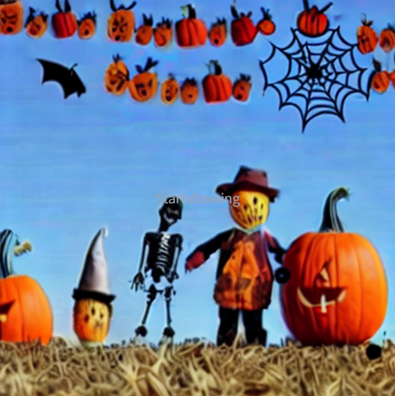}  \\

User & I added a skeleton, a ghost and a spider web to my Halloween poster like you suggested. I am pretty happy with my final product. What do you think? \\
Assistant & 
    I think your Halloween poster looks great! The addition of the skeleton, ghost, and spider web adds a spooky and festive touch to the poster. The combination of the pumpkins, skeleton, and spider web creates a visually appealing and fun design that captures the essence of Halloween. Well done!

\end{tabular}
}
\end{AIbox}
\vspace{-2mm}
\caption{
  Graphic Design for a Halloween Poster (3/3).  Refer to the scenario discussions in Section~\ref{sec:graphics_design_halloween}.
} \label{fig:graphics_design_halloween_3}
\vspace{-0mm}
\end{figure}

Please see the user-assistant interaction in Figure~\ref{fig:graphics_design_halloween_1}, ~\ref{fig:graphics_design_halloween_2} and 
~\ref{fig:graphics_design_halloween_3}. Designing visually appealing posters and gift cards for occasions like Halloween demands imaginative concepts and striking aesthetics. For instance, when crafting a Halloween poster, the user may request the AI assistant to offer a range of ideas, and then choose one to be transformed into an image. To refine the generated image, the user may seek further suggestions, such as incorporating a bat, swapping the scarecrow with a ghost, removing a smaller scarecrow, adding a skeleton, and substituting pumpkins with a spider web. After making these adjustments, the user can request feedback, and the assistant affirms that the design effectively captures the spirit of Halloween.

\subsection{Fashion Design for Kid's Clothing}
\label{sec:kid_cloth_picture_design}
\begin{figure}[!ht]
\begin{AIbox}{Fashion Design for Kid's Clothing (1/2)}

\scalebox{0.83}{
\begin{tabular}{p{1.3cm} p{13.2cm}  }
&  \includegraphics[height=6.0cm]{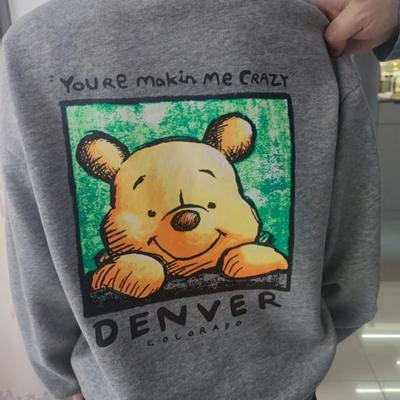} \\

User & what is written in the image? \\

Assistant & The image features a sweatshirt with the words "You're making me crazy" written on it. \\

User & 这是什么意思？ \\

Assistant & 这句话的意思是：你让我觉得疯狂。 \\

User & \textcolor{magenta}{(Inpaint New Objects) [Draw boxes for \texttt{Grounding Instruction} = ``blue hat; sun glasses''] [Generate]}  \\

Assistant & 
 \includegraphics[height=6.0cm]{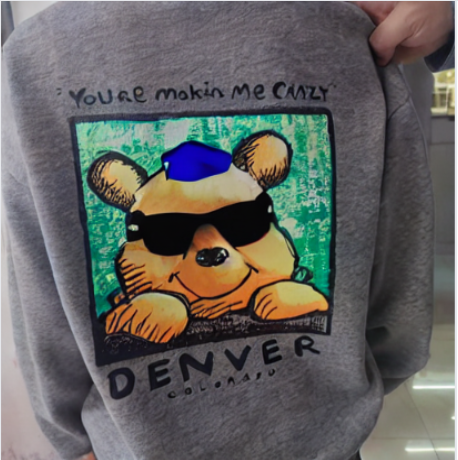} \\

User & how do you think about the picture design as a kid cloth? \\

Assistant & When designing a picture for a kid's cloth, it's important to consider the age range and interests of the child who will be wearing it. For younger children, simple shapes, bright colors, and characters from popular children's shows or movies can be effective. For older children, more complex designs, patterns, and popular trends in fashion can be considered. It's also important to ensure that the design is age-appropriate and does not contain any inappropriate or offensive imagery. Additionally, the design should be durable and able to withstand washing and wear. In the case of the image, the design features a cartoon bear wearing sunglasses and a hat, which could be appealing to children who enjoy animals or have a sense of humor. The sweatshirt is also personalized with the name "Denver" on the back, which could be a way to show pride in the child's hometown or a place they enjoy visiting. \\

User & \textcolor{magenta}{(Remove and Change Objects) [\texttt{Referral Text} = ``bear and glasses'', visual prompt by drawing on ``hand'', then remove them by dragging the segmentation masks out of the image] [Generate]}  \\

Assistant & 
 \includegraphics[height=6.0cm]{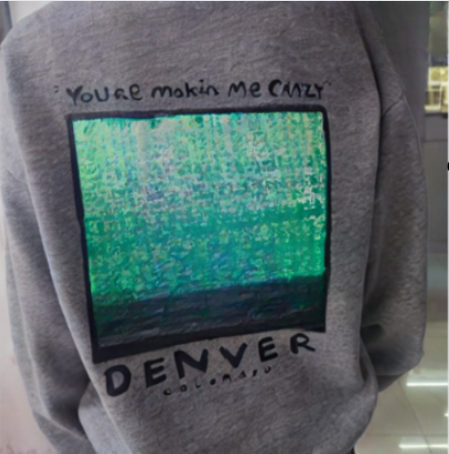} \\

\end{tabular}
}

\end{AIbox}
\vspace{-2mm}
\caption{
 Fashion Design for Kid's Clothing (1/2).  Refer to the scenario discussions in Section~\ref{sec:kid_cloth_picture_design}.
} \label{fig:kid_cloth_picture_design_1}
\vspace{-0mm}
\end{figure}

\begin{figure}[!ht]
\begin{AIbox}{Fashion Design for Kid's Clothing (2/2)}

\scalebox{0.83}{
\begin{tabular}{p{1.3cm} p{13.2cm}  }

User & \textcolor{magenta}{(Inpaint New Objects) [Draw boxes for \texttt{Language Instruction} = ``boat; lake; snow mountain; tent''] [Generate]}  \\
 & 
 \includegraphics[height=6.0cm]{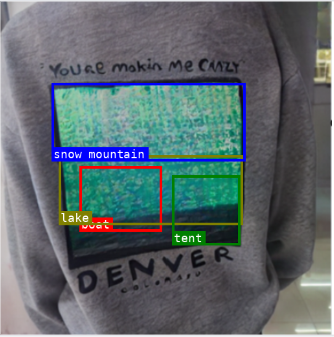} \\

Assistant &   \includegraphics[height=6.0cm]{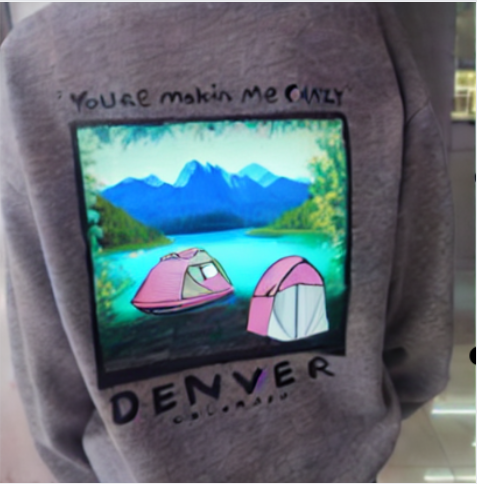} \\

User &  how do think about the current picture design for a kid cloth? \\
Assistant & 
 The current picture design for the kid's cloth is quite appealing and imaginative. The scene of two tents on a lake, surrounded by mountains, creates a sense of adventure and outdoor exploration, which can be both entertaining and inspiring for a child. The vibrant colors and detailed illustration make the design stand out and can spark a child's imagination, making it a fun and creative choice for a kid's cloth. The design is visually engaging and can capture a child's attention, making it an attractive option for a child's clothing. Overall, the current picture design is a great choice for a kid's cloth, as it combines both aesthetics and imagination, creating a visually appealing and engaging piece of clothing. \\

\end{tabular}
}

\end{AIbox}
\vspace{-2mm}
\caption{
 Fashion Design for Kid's Clothing (2/2).  Refer to the scenario discussions in Section~\ref{sec:kid_cloth_picture_design}.
} \label{fig:kid_cloth_picture_design_2}
\vspace{-0mm}
\end{figure}

Please see the user-assistant interaction in Figure~\ref{fig:kid_cloth_picture_design_1} and  ~\ref{fig:kid_cloth_picture_design_2}. Imagine one day when a user sees his Chinese nephew's sweatshirt with designed English text and a picture on the back, and wants to personalize the design while the sweatshirt is being worn. To first show his nephew what is written on the sweatshirt and the meaning of the text. The user asks the related questions, and the assistant is able to correctly answer both. This simple application can be widely used to recognize text on various clothes on the Chinese market, where most users have less knowledge about designed English text. 

With \shortname{}, the user can further edit the picture by following his nephew's personalized requests to add a blue hat and sun glasses. The assistant can make comments on the new picture, saying that `` the design features a cartoon bear wearing sunglasses and a hat, which could be appealing to children who enjoy animals or have a sense of humor.'' This encouraging comment can boost the kid's confidence in his design skills. 

The assistant also mentions the city ``Denver'' in the comments, which the assistant guesses that it could be the child's hometown or a place they enjoy visiting. Following the comments, the user wants to design a new picture with a representative natural scene of Denver. To this end, the user first removes the bear and only leaves the background; after that, the user creates a new scene by specifying the spatial layout of objects ``lake; boat; tent; snow mountains''. With the new image, the assistant believes that it is a great choice for a kid's clothing, as it combines both aesthetics and imagination.

\subsection{Food Preparation}
\label{sec:dinner_preparation}
\begin{figure}[!ht]
\begin{AIbox}{Dinner Preparation}

\scalebox{0.83}{
\begin{tabular}{p{1.3cm} p{13.6cm}  }
&  \includegraphics[height=6.0cm]{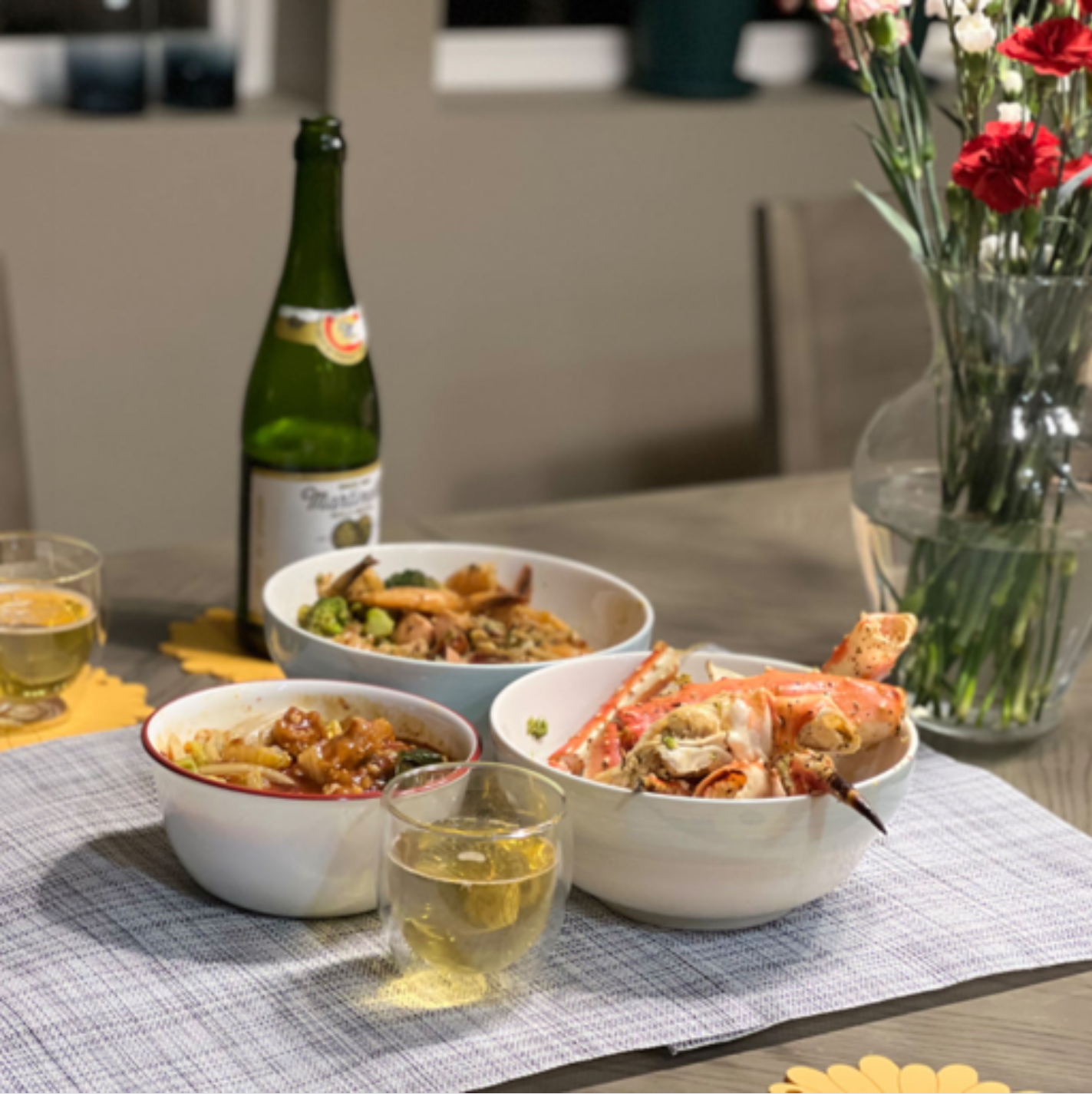} \\

User & I just prepared a dinner for my girlfriend, how do you think of it? is there anything l should add or remove to improve the dinner? \\

Assistant & The dinner you prepared looks delicious and well-presented. The table is set with a variety of food, including bowls of Asian cuisine, a bottle of wine, and wine glasses. There are also vases with flowers, adding a touch of elegance to the setting. The table is set with chairs, and the overall presentation is inviting.

As for improvements, it would be great to have some side dishes or salads to complement the main courses. You could also consider adding some dipping sauces or condiments to enhance the flavors of the dishes. Additionally, you could add some soft lighting or candles to create a more intimate and cozy atmosphere. Finally, you could play some background music to set the mood for a romantic dinner. \\

User & \textcolor{magenta}{(Inpaint New Objects) [Draw boxes for \texttt{Grounding Instruction} = ``salads; candles''] [Generate]}  \\

Assistant & 
 \includegraphics[height=6.0cm]{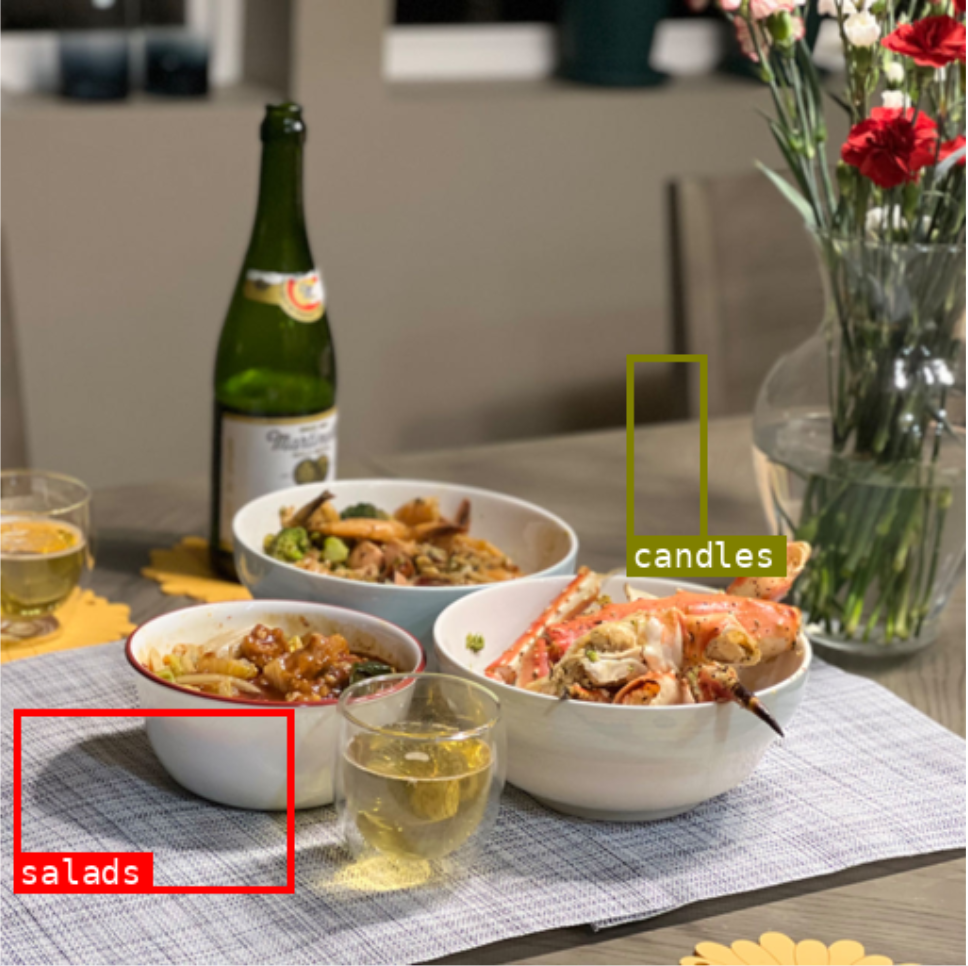} 
 \includegraphics[height=6.0cm]{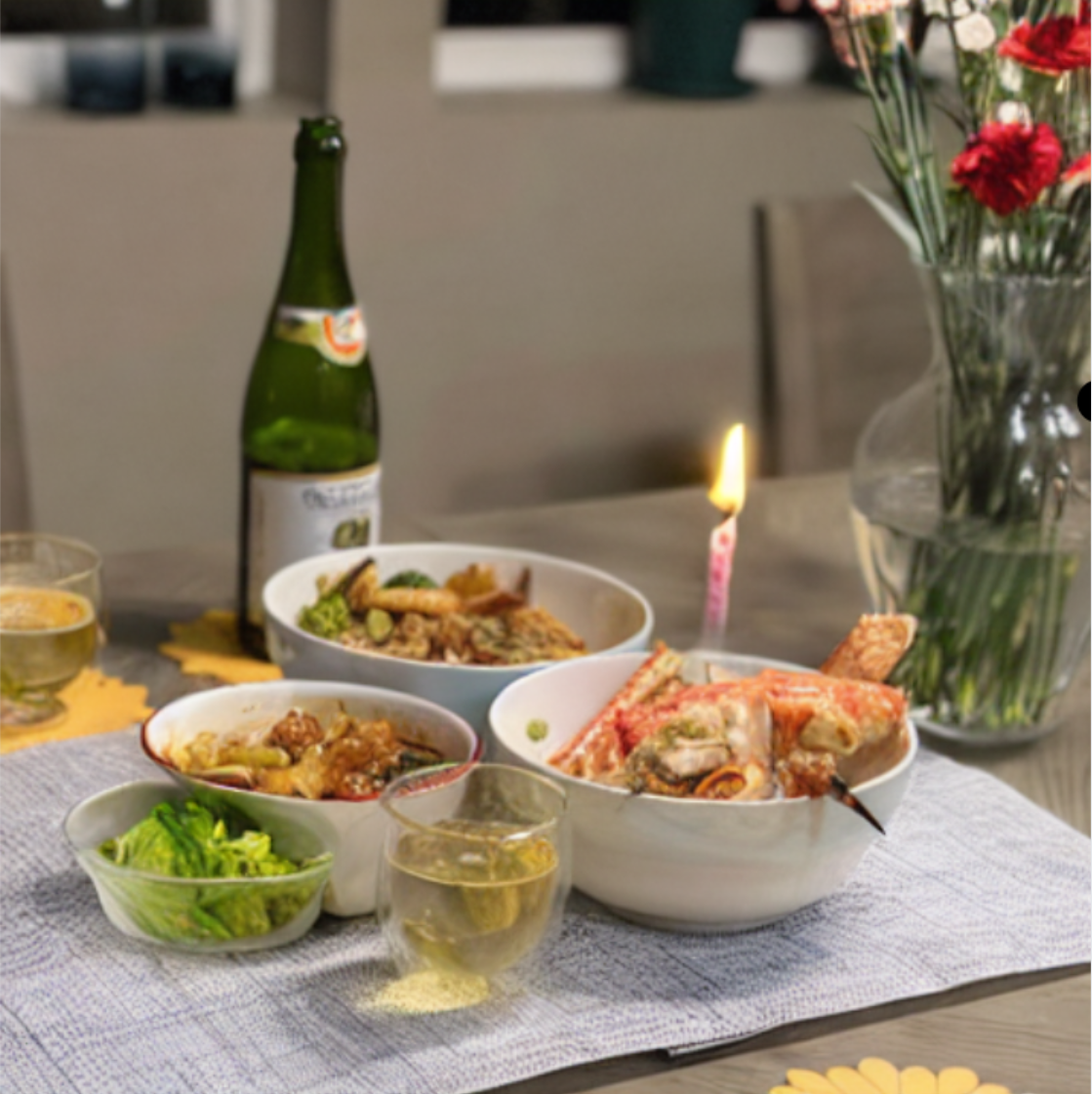}
  \\
User & salads and candles are added now. is the current dinner good?\\

Assistant & The current dinner setup appears to be quite appealing and well-prepared. The table is set with a variety of dishes, including bowls of food, a bottle of wine, and a lit candle. The presence of salads and candles adds to the ambiance and makes the dinner more enjoyable. The combination of food, wine, and candlelight creates a warm and inviting atmosphere, which is ideal for a pleasant dining experience. \\

User &  This image is the home dinner I made; can you give dating behavior tips for this particular dinner?\\
Assistant & 
Here are some dating behavior tips for a home-cooked dinner like the one in the image:

Prepare the meal: Show your date that you put effort into making the meal by preparing it yourself. This can be a fun and creative way to spend time together.
Set the mood: Create a cozy and intimate atmosphere at home. Light some candles, play some soft music, and set the table with nice plates and cutlery.
Communicate: Use this opportunity to catch up and have a meaningful conversation with your date. Ask them about their day, share your thoughts and feelings, and engage in deep conversation.
Be attentive: Pay attention to your date's needs and preferences. Offer to refill their drink, ask if they would like more food, and make sure they are comfortable.
Show appreciation: Express your gratitude for your date's company and the meal. Compliment their outfit or tell them how much you enjoyed the food.
Be yourself: Be authentic and relaxed. Don't try to impress your date with a fancy meal or extravagant gesture. Instead, focus on spending quality time together and enjoying each other's company.
Have fun: Finally, have fun and enjoy the evening! Laugh, share stories, and create memories together.
In the image, there are several bowls of food, a bottle of wine, and a lit candle on the table. This suggests that the meal is a special occasion, and the couple is enjoying each other's company in a cozy and intimate setting. \\

\end{tabular}
}

\end{AIbox}
\vspace{-2mm}
\caption{
 Dinner preparation for a date.  Refer to the scenario discussions in Section~\ref{sec:dinner_preparation}.
} \label{fig:dinner_preparation}
\vspace{-0mm}
\end{figure}

\paragraph{Dinner Preparation for a Romantic Date}
Please see the user-assistant interaction in Figure~\ref{fig:dinner_preparation}. 
Individuals often enjoy capturing photos of appetizing meals they consume. When planning a romantic dinner, individuals typically dedicate time and consideration to crafting the perfect dish, complemented by wine and flowers. Nonetheless, they may still feel apprehensive about whether the dinner is adequately prepared or if any enhancements could be made. 

In this situation, we suggest utilizing \shortname{} to obtain valuable advice and recommendations. The assistant expresses enthusiasm about the dinner while also offering specific ideas, such as incorporating salads to enhance the main dishes and using candles to establish a warm and intimate ambiance. By implementing these suggestions, the user can modify the image to develop various virtual dinner options and submit them to the assistant for evaluation. Once the ideal solution is determined and receives positive feedback from the assistant, the user can also request guidance on appropriate dating etiquette for this specific dinner.

\begin{figure}[!ht]
\begin{AIbox}{Recipe (1/2)}
\scalebox{0.83}{
\begin{tabular}{p{1.3cm} p{13.2cm}  }

& \includegraphics[height=6.0cm]{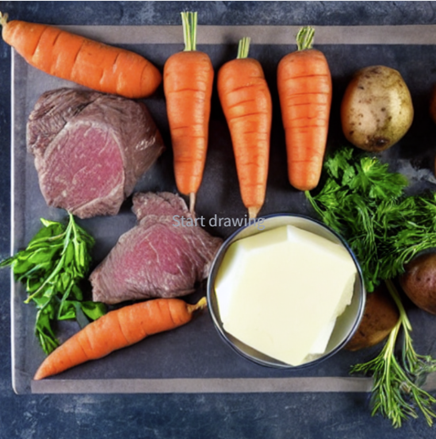} \\
 
User & I have those products except the greens and would like to cook a meal. What could I cook? Could you please give me the recipe. \\
Assistant & 
    Certainly! You can make a delicious roast beef and vegetable dinner using the ingredients you have. Here's a simple recipe to get you started:\newline
    Ingredients:
    \begin{itemize}
        \item 1 lb roast beef (sliced)
        \item 4-6 medium carrots (peeled and chopped)
        \item 2-3 medium potatoes (peeled and chopped)
        \item 1/2 cup butter
        \item Salt and pepper to taste
        \item Optional: 1 tsp dried herbs (such as rosemary, thyme, or parsley)
    \end{itemize}
    Instructions:
    \begin{enumerate}
        \item Preheat your oven to 350°F (175°C).
        \item In a large roasting pan, place the sliced roast beef. Season with salt and pepper, and add the optional dried herbs if desired.
        \item Arrange the chopped carrots and potatoes around the roast beef, ensuring they are evenly distributed.
        \item Drizzle the butter over the vegetables and roast beef, making sure it is evenly coated.
        \item Cover the roasting pan with aluminum foil and place it in the preheated oven.
        \item Roast the beef and vegetables for about 1 hour, or until the meat is cooked through and the vegetables are tender.
        \item Remove the roasting pan from the oven and carefully remove the foil.
        \item Use a spatula or serving spoon to serve the roast beef and vegetables onto individual plates.
        \item Enjoy your delicious roast beef and vegetable dinner!        
    \end{enumerate}
    This recipe is a simple and hearty meal that can be enjoyed by the whole family. You can also customize the vegetables and herbs to suit your preferences, and adjust the cooking time as needed to achieve the desired level of doneness.\\
    
    User & \textcolor{magenta}{(Remove or Change Objects) [\texttt{Referral Text} = "butter"][Segment][\texttt{Grounding Instruction} = "rice"][Generate]} \\
    Assistant &  \includegraphics[height=4.0cm]{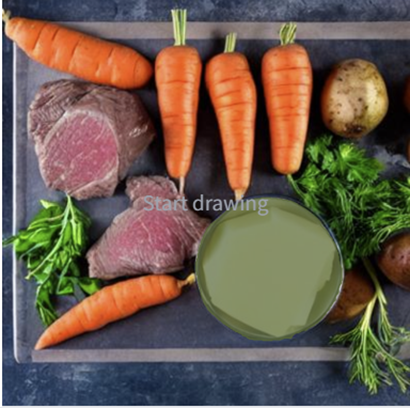}  \includegraphics[height=4.0cm]{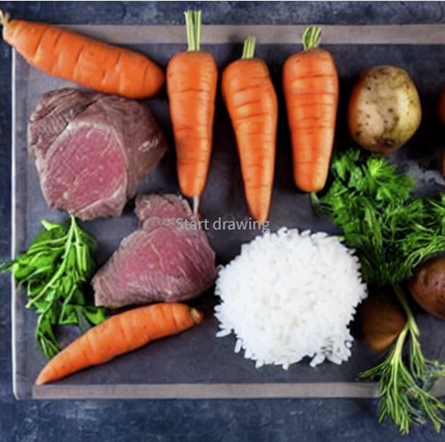} \\ 

 \end{tabular}
}
\end{AIbox}
\vspace{-2mm}
\caption{
 Recipe (1/2).  Refer to the scenario discussions in Section~\ref{sec:dinner_preparation}.
} \label{fig:dinner_preparation2_1}
\vspace{-0mm}
\end{figure}

\paragraph{Food Preparation Recipe}
Please see the user-assistant interaction in  Figure~\ref{fig:dinner_preparation2_1} and~\ref{fig:dinner_preparation2_2}. In another example involving meal preparation using available ingredients, the user may inquire about the necessary ingredients and cooking instructions. The user might also decide to change some elements, such as replacing butter with rice. Upon asking the same question again, updated ingredients and instruction lists will be provided with the revised cooking tips.

\subsection{Visual Content Creation and Story Telling}
\label{sec:visual_story}
\begin{figure}[!ht]
\begin{AIbox}{Visual Content Creation and Story Telling}

\scalebox{0.81}{
\begin{tabular}{p{1.3cm} p{13.8cm}  }

&  \includegraphics[height=6.0cm]{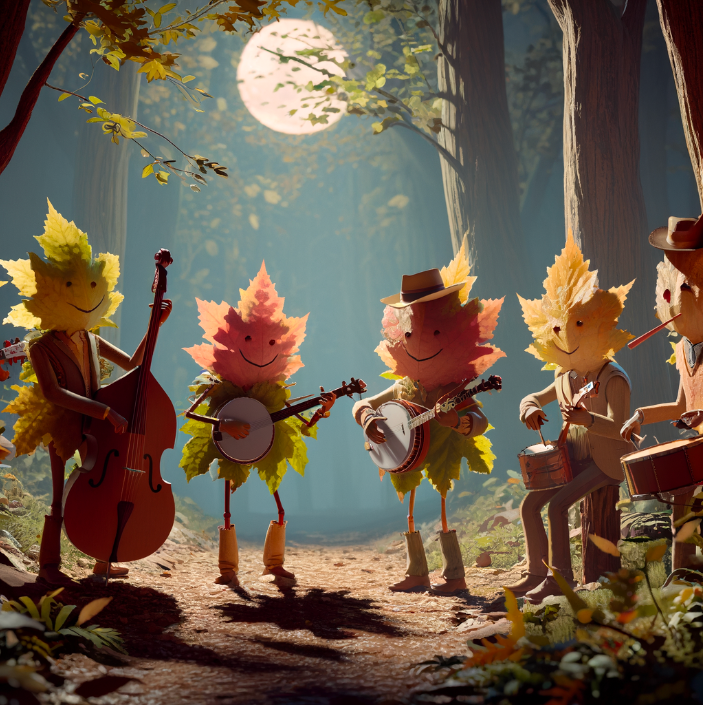} \\

User & can you describe the image? \\

Assistant & The image features a group of five cartoon characters, each dressed as a different type of tree, playing musical instruments in a forest setting. They are positioned in a line, with one character on the left, two in the middle, and two on the right side of the scene. The characters are playing various instruments, including a violin, a guitar, and a banjo.

The forest setting is enhanced by the presence of trees in the background, with one large tree on the left side and another on the right side of the image. The scene appears to be a lively and whimsical representation of a musical performance in a natural environment. \\

User & please write a story based on the image? \\

Assistant & Once upon a time, in a magical forest, a group of five tree-themed characters gathered to perform a musical concert. Each character was dressed in a unique costume representing their respective tree species, and they played various musical instruments with great enthusiasm.

The first character, a tall and slender tree, played the violin with grace and precision. The second character, a sturdy and broad tree, strummed the guitar with a powerful and rhythmic beat. The third character, a delicate and elegant tree, skillfully plucked the banjo, adding a touch of melody to the performance. The fourth character, a small and twisted tree, played the drums with a lively and energetic flair. Lastly, the fifth character, a short and bushy tree, danced and sang, captivating the audience with their charming presence.

As the concert progressed, the forest came alive with the music, and the audience of animals and fairies gathered around to enjoy the performance. The moon, which was full and bright, illuminated the scene, casting a warm and magical glow over the entire forest.

The concert was a great success, and the tree-themed characters became known as the most talented and entertaining musicians in the land. They continued to perform together, spreading joy and happiness through their music, and the forest became a place of wonder and enchantment for all who visited. \\

User & if necessary, please provide suggestion to edit the image for a more whimsical story? \\

Assistant & To make the image more whimsical, one could consider adding more playful and imaginative elements to the scene. For example, the characters could be posed in a more exaggerated and humorous manner, with oversized instruments or exaggerated facial expressions. The forest setting could be further enhanced by adding more colorful and fantastical elements, such as glowing mushrooms, sparkling flowers, or even a magical waterfall.

Additionally, the characters could be interacting with each other in a more playful and lighthearted manner, such as dancing, laughing, or even engaging in a friendly competition to see who can play the best instrument. The overall atmosphere of the image could be further enhanced by adding more whimsical details, such as a flying bird or a playful squirrel, to create a more imaginative and delightful scene. \\

User & \textcolor{magenta}{(Inpaint New Objects) [Draw boxes for \texttt{Grounding Instruction} = ``glowing mushrooms; drums'' ] [Generate]}  \\

Assistant &  
 \includegraphics[height=6.0cm]{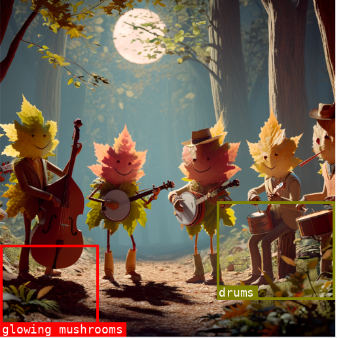} 
\includegraphics[height=6.0cm]{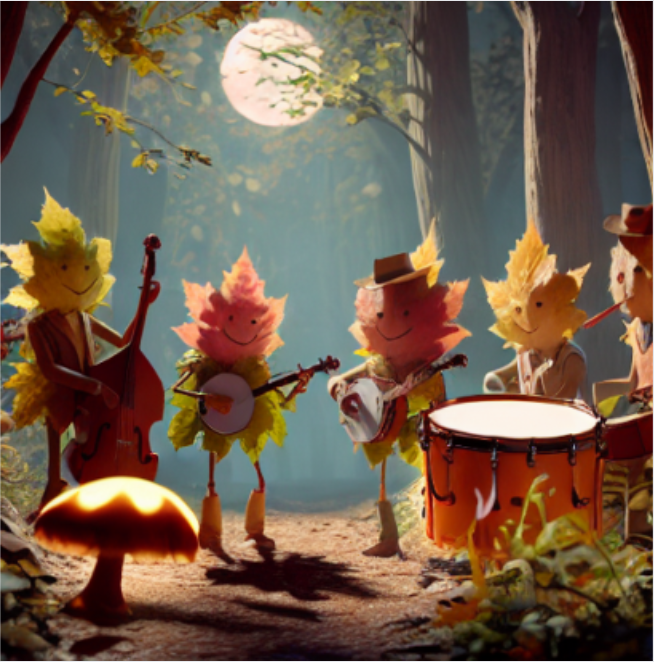} \\

\end{tabular}
}

\end{AIbox}
\vspace{-2mm}
\caption{Visual content creation and story telling.  Refer to the scenario discussions in Section~\ref{sec:visual_story}.
} \label{fig:visual_story}
\vspace{-0mm}
\end{figure}

Please see the user-assistant interaction in Figure~\ref{fig:visual_story}. 
The process of visual storytelling usually demands creativity and considerable time investment, as it involves the development of both compelling images and imaginative text. Occasionally, adjustments to the visuals may be necessary to ensure they are in harmony with the overall scene. In Figure~\ref{fig:visual_story}, \shortname{} is able to provide detailed descriptions and a magical story for kids. 

The user may ask for possible edits on the image for a more whimsical story. Several more playful and imaginative elements are suggested, including glowing mushrooms and oversize instruments. By following the idea, the user inpaints the new objects of intended spatial configuration in the image, \eg a growing mushroom in front of the first character on the left, and an oversize drum for the fourth character who is playing a drum.

\subsection{Education}
\label{sec:education}
\begin{figure}[!ht]
\begin{AIbox}{Scientific Education (1/2)}
\scalebox{0.83}{
\begin{tabular}{p{1.3cm} p{13.2cm}  }

User & \textcolor{magenta}{(Generate New Image) [Draw box for \texttt{Grounding Instruction} = "the space needle; dinosaur"][\texttt{Language Instruction} = "the space needle and dinosaur"] [Generate]} \\
Assistant & \includegraphics[height=4.0cm]{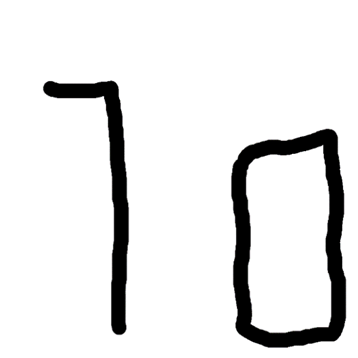} \includegraphics[height=4.0cm]{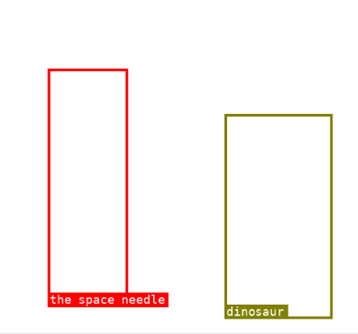}  \includegraphics[height=4.0cm]{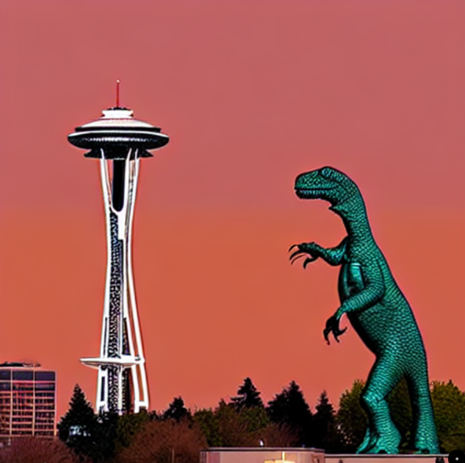} \\

User & Can you tell me what is in this picture? \\
Assistant & 
The image features a large green dinosaur statue standing next to a tall tower, possibly the Space Needle. The dinosaur appears to be a T-Rex, and it is positioned in front of the tower, creating an interesting contrast between the two structures. The scene is set against a backdrop of a pink sky, adding a unique and vibrant atmosphere to the image. \\

User &  What is a T-rex? \\
Assistant &  A T-Rex, short for Tyrannosaurus Rex, is a large carnivorous dinosaur that lived during the Late Cretaceous period, around 68 to 65 million years ago. It is one of the most well-known and iconic dinosaurs due to its massive size, powerful jaws, and distinctive appearance. T-Rex had a small head, large eyes, and a long, sturdy tail. It was a dominant predator in its ecosystem, feeding on a variety of prey, including other dinosaurs, such as Triceratops and Edmontosaurus.
T-Rex is often depicted in popular culture, including movies, television shows, and video games, as a fearsome and relentless predator. However, in reality, it is believed that T-Rex had a relatively slow and lumbering gait, and it may have been more of a scavenger than an active hunter. Despite its fearsome reputation, T-Rex was not the largest land predator that ever existed; that title belongs to the Spinosaurus, a dinosaur from the same time period.\\

User &  Why is the sky pink? \\
Assistant & 
The sky is pink due to the presence of clouds in the atmosphere, which are reflecting the sunlight. This phenomenon occurs when sunlight passes through the Earth's atmosphere and interacts with the molecules in the air, causing them to scatter the light. The scattering process is more pronounced for shorter wavelengths, such as blue and violet, than for longer wavelengths, like red and orange. As a result, the sky appears pink during sunset or sunrise, when the sun is low in the sky and the light has to travel through more of the Earth's atmosphere.\\

User &  If I want this picture to represent midday, not during sunset or sunrise, what color should the sky be? \\
Assistant & 
If you want the picture to represent midday, the sky should be blue. During midday, the sun is at its highest point in the sky, and the light has to travel through the least amount of the Earth's atmosphere. This results in the scattering of shorter wavelengths, such as blue and violet, being less pronounced, allowing the longer wavelengths, like red and orange, to dominate the sky's color. In this case, the sky would appear blue, providing a more natural and accurate representation of the time of day. \\

User & \textcolor{magenta}{(Change and Remove Objects) [\texttt{Referral Text} = "T-rex"][Segment]} \\
Assistant & \includegraphics[height=6.0cm]{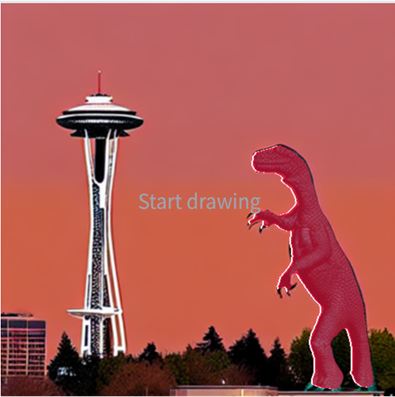} \\
\end{tabular}
}
\end{AIbox}
\vspace{-2mm}
\caption{
Scientific Education (1/2).  Refer to the scenario discussions in Section~\ref{sec:education}.
} \label{fig:education_dinosaur_1}
\vspace{-0mm}
\end{figure}

\begin{figure}[!ht]
\begin{AIbox}{Scientific Education (2/2)}
\scalebox{0.83}{
\begin{tabular}{p{1.3cm} p{13.2cm}  }

User & I want to replace the dinosaur with something else. Any suggestions? \\
Assistant & 
    Certainly! If you would like to replace the dinosaur with something else, you could consider the following options:
    1.	A giant robot or mech: This could be a futuristic or retro-inspired design, adding a sense of wonder and excitement to the scene.
    2.	A massive sculpture of a famous historical figure: This could be a person from history, mythology, or even a fictional character, such as a superhero or a popular movie character.
    3.	A giant, intricate artwork or mural: This could be a colorful, abstract, or representational piece that adds visual interest and artistic flair to the area.
    4.	A large, interactive exhibit or installation: This could be an immersive experience that engages visitors, such as a virtual reality exhibit, an interactive art piece, or a themed playground.
    Ultimately, the choice of replacing the dinosaur with something else depends on the context, the desired atmosphere, and the preferences of the local community or visitors. \\

User & \textcolor{magenta}{(Change and Remove Objects) [\texttt{Grounding Instruction} =  "a giant robot"][Generate]} \\
Assistant & \includegraphics[height=6.0cm]{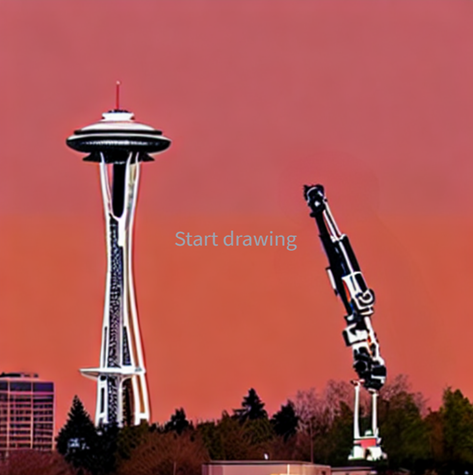} \\
User & What kind of robot is that? \\
Assistant & 
    The image features a large robotic crane, which is a type of construction equipment used for lifting and moving heavy objects.\\

User & \textcolor{magenta}{(Change and Remove Objects) [Generate]} \\
Assistant & \includegraphics[height=6.0cm]{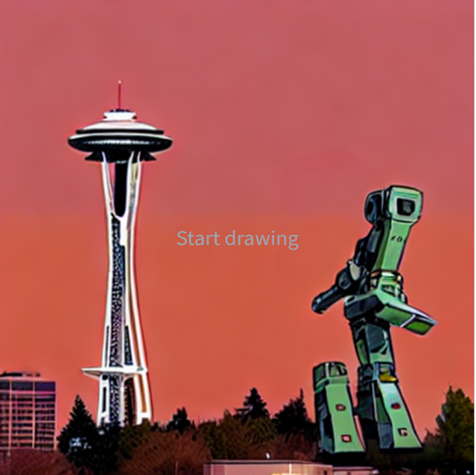} \\
User & What kind of robot is that? \\
Assistant &  The robot in the image is a large, green, metallic robot with a humanoid shape. \\
User & What is this type of robot used for? \\
Assistant & The large green robot is a mechanical robot, which is typically used for tasks that require strength, precision, or the ability to perform repetitive actions. These robots can be found in various industries, such as manufacturing, construction, and even entertainment. \\

User & Thank you so much for the education. \\
Assistant & You're welcome! I'm glad I could help. If you have any more questions, feel free to ask. \\

\end{tabular}
}
\end{AIbox}
\vspace{-2mm}
\caption{
Scientific Education. (2/2) Refer to the scenario discussions in Section~\ref{sec:education}.
} \label{fig:education_dinosaur_2}
\vspace{-0mm}
\end{figure}

\paragraph{Scientific Education.}
Please see the user-assistant interaction in Figure~\ref{fig:education_dinosaur_1} and ~\ref{fig:education_dinosaur_2}.
In order to engage children in learning scientific concepts, it is effective to present the information using visual imagery and familiar themes, such as cartoon characters. For instance, a child living in the Seattle area may be drawn to an image featuring the Space Needle and a dinosaur set against a pink sky, creating a lively scene. Such an image would capture the child's interest due to its recognizable elements. The child might then inquire about the T-rex, the reason behind the pink sky, the color of the sky during midday, and why it changes. Additionally, the T-rex could be replaced with various robots, prompting the child to ask about their functions and concepts. By using familiar and visually appealing elements, learning about science can become an enjoyable experience for children.

\begin{figure}[!ht]
\begin{AIbox}{Education: Cartoon Reading.}
\scalebox{0.83}{
\begin{tabular}{p{1.3cm} p{13.2cm}  }

User & \includegraphics[height=6.0cm]{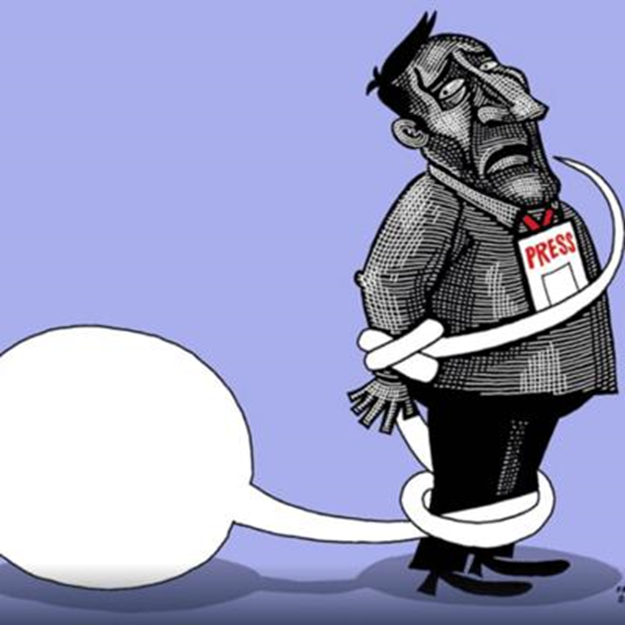} \\
User & What does this editorial cartoon mean?\\
Assistant & 
    The editorial cartoon depicts a man in a suit, who appears to be a reporter or a news anchor, being tied up with a ball and chain. This image symbolizes the idea of being restricted or limited in one's freedom of expression or movement, possibly due to censorship or other forms of control. The man's profession as a reporter or news anchor suggests that the cartoon might be commenting on the challenges faced by journalists in certain situations, where they may be subjected to restrictions or limitations on their ability to report the news freely and accurately. The image serves as a metaphor for the challenges faced by journalists in their pursuit of truth and the importance of protecting their freedom of expression.\\

User & \textcolor{magenta}{(Change and Remove Objects) [Visual prompt by stroking on on "tie"][Segment][Drag to remove][Generate]} \\
Assistant & \includegraphics[height=4.0cm]{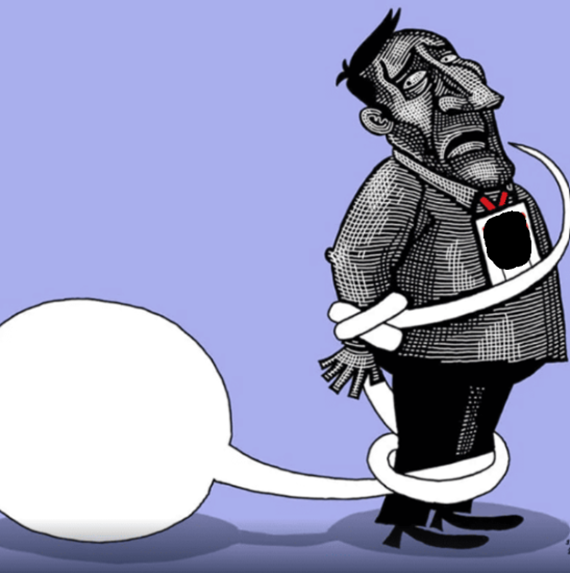} \includegraphics[height=4.0cm]{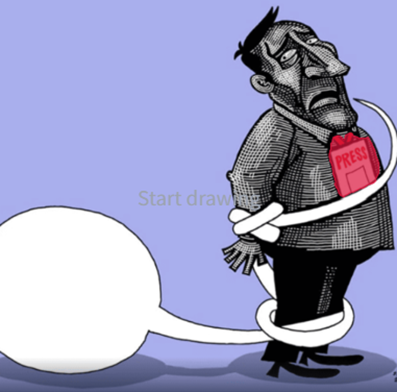} \includegraphics[height=4.0cm]{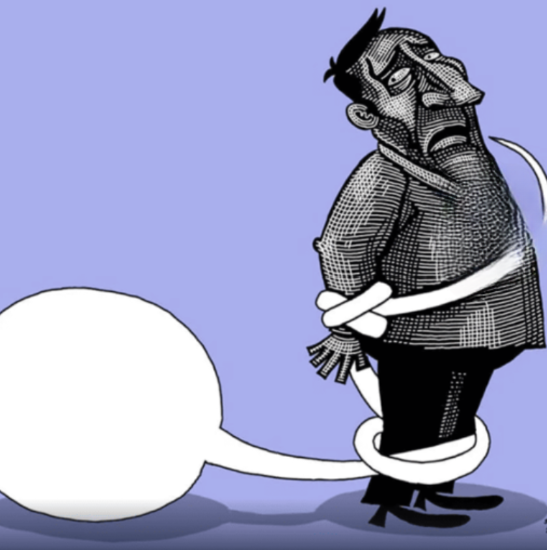}\\
User & What does this editorial cartoon mean? \\
Assistant & 
    The editorial cartoon features a man in a suit and tie, who is being pulled by a giant rubber band. The rubber band is attached to the man's waist, and it appears to be stretched to its limit. The image is a black and white drawing, which adds to the dramatic effect of the scene.The meaning of this editorial cartoon could be interpreted in various ways, depending on the context in which it was created. It could symbolize the challenges or burdens that the man might be facing, such as financial or professional responsibilities. The rubber band could represent the pressure or stress that the man is experiencing, and the fact that it is stretched to its limit might indicate that he is struggling to cope with these challenges. Alternatively, the image could also be interpreted as a humorous representation of the man's size or the consequences of overeating. The cartoon might be commenting on the man's weight or the societal expectations of professionalism and appearance. In any case, the cartoon is a powerful visual representation of the challenges and pressures that the man might be facing, and it encourages the viewer to reflect on the underlying message and its relevance to their own experiences or the broader context in which the cartoon was created.\\

\end{tabular}
}
\end{AIbox}
\vspace{-2mm}
\caption{
Education: Cartoon Reading. Refer to the scenario discussions in Section~\ref{sec:education}.
} \label{fig:education_cartoon}
\vspace{-0mm}
\end{figure}

\paragraph{Cartoon Reading Education.}
Please see the user-assistant interaction in Figure~\ref{fig:education_cartoon}. 
To improve one's ability to interpret cartoons, it is crucial to understand that different details within an image can communicate various meanings. For example, when examining an editorial cartoon featuring a man wearing clothing labeled "PRESS" and bound by a ball and chain, the assistant might explain that the image metaphorically represents the obstacles journalists encounter in their quest for truth and the significance of safeguarding their freedom of expression. However, if the "PRESS" label is removed and the same question is asked, the responses could vary greatly. The cartoon may still convey a strong visual message about the challenges and pressures the man faces, but the interpretations can differ considerably without the context provided by the "PRESS" label.

\subsection{Interior Design}
\label{sec:interior_design}
\begin{figure}[!ht]
\begin{AIbox}{Interior Design: Large Living Room (1/2)}
\scalebox{0.83}{
\begin{tabular}{p{1.3cm} p{13.2cm}  }

User & \textcolor{magenta}{(Generate New Image) [\texttt{Language Instruction} = "Living room with minimum furniture and nice lighting"][Generate]} \\
Assistant & \includegraphics[height=6.0cm]{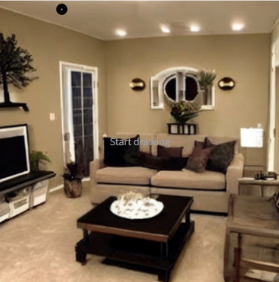} \\
User & What do you think about this room. How could I improve it to make it more modern? \\
Assistant &
    The room appears to be a cozy living room with a couch, a coffee table, a TV, and a potted plant. To make it more modern, you could consider updating the furniture and decor to reflect contemporary styles. For example, you could replace the couch and coffee table with sleek, minimalist designs, and add a stylish rug to the floor. Additionally, you could update the TV to a slim, flat-screen model and consider adding some modern artwork or decorative items on the walls. To enhance the room's ambiance, you could also introduce more potted plants or greenery, and consider using a combination of neutral and warm colors for the walls, curtains, and other textiles. Finally, adding some ambient lighting, such as floor lamps or table lamps, can create a more inviting and modern atmosphere in the living room.\\

User & \textcolor{magenta}{(Remove or Change Object) [\texttt{Referral text} = "sofa"] [Segment][\texttt{Grounding Instruction} =  "Contemporary light sofa"] [Generate]} \\
Assistant & \includegraphics[height=6.0cm]{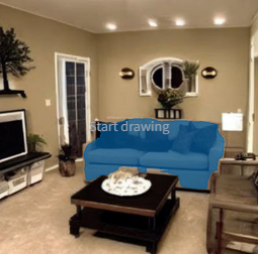} \includegraphics[height=6.0cm]{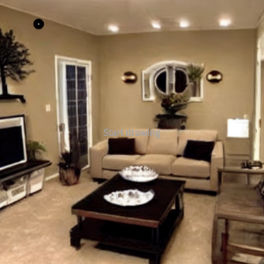}\\

User & \textcolor{magenta}{(Remove or Change Object) [\texttt{Referral text} = "coffee table"] [Segment][\texttt{Grounding Instruction} =  "contemporary coffee table"] [Generate]} \\
Assistant & \includegraphics[height=6.0cm]{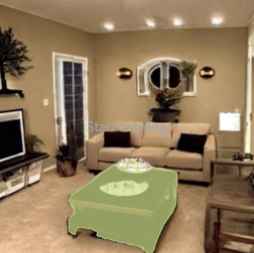} \includegraphics[height=6.0cm]{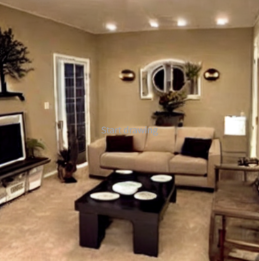}\\

\end{tabular}
}
\end{AIbox}
\vspace{-2mm}
\caption{
Interior Design: Large Living Room (1/2). Refer to the scenario discussions in Section~\ref{sec:interior_design}.
} \label{fig:design_large_room_1}
\vspace{-0mm}
\end{figure}

\begin{figure}[!ht]
\begin{AIbox}{Interior Design: Large Living Room (2/2)}
\scalebox{0.83}{
\begin{tabular}{p{1.3cm} p{13.2cm}  }

User & \textcolor{magenta}{(Remove or Change Object) [Visual prompt by stroking on chair] [Segment][\texttt{Grounding Instruction} = "Potted plant"] [Generate]} \\
Assistant & \includegraphics[height=4.0cm]{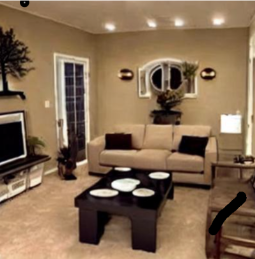} \includegraphics[height=4.0cm]{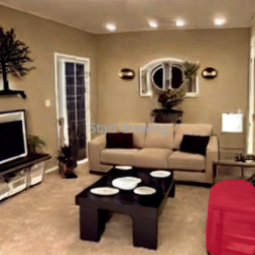} \includegraphics[height=4.0cm]{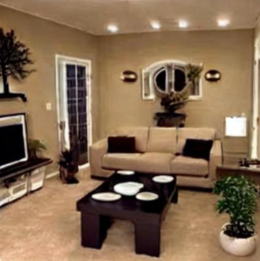} \\

User & \textcolor{magenta}{(Inpaint New Objects) [Draw box for \texttt{[Grounding Instruction} = "Art work by Gustav Klimt"] [Generate]} \\
Assistant & \includegraphics[height=6.0cm]{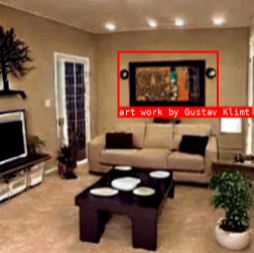} \includegraphics[height=6.0cm]{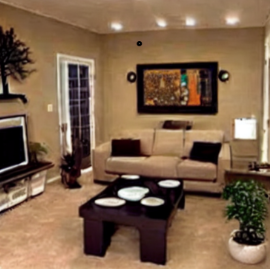} \\

User & I have replaced the couch, coffee table, added potted plant and added modern painting to this room. I started to like it. What do you think? \\
Assistant &
    It sounds like you have made some great improvements to the living room, and the addition of the modern painting, sleek furniture, and contemporary decor has likely transformed the space into a more modern and stylish environment. The potted plant and rug also add to the overall aesthetic, creating a warm and inviting atmosphere. It's always great to hear that someone is pleased with their living space, as it reflects their personal taste and preferences. The combination of these elements has likely resulted in a more comfortable and visually appealing living room that you can enjoy spending time in.

\end{tabular}
}
\end{AIbox}
\vspace{-2mm}
\caption{
Interior Design: Large Living Room (2/2).  Refer to the scenario discussions in Section~\ref{sec:interior_design}.
} \label{fig:design_large_room_2}
\vspace{-0mm}
\end{figure}

\paragraph{Interior Design: Large Living Room.}
Please see the user-assistant interaction in Figure~\ref{fig:design_large_room_1} and~\ref{fig:design_large_room_2}.
Interior design can be progressively enhanced through a trial-and-error process. Initially, the user creates an image based on a text description prompt and seeks advice on how to modernize the design. Suggestions might include incorporating minimalist elements, adopting contemporary styles, adding potted plants, and displaying modern art on the walls. The user can then modify the design by segmenting objects and making adjustments using stroke and text prompts. As a result, these new elements come together to create a more comfortable and updated living room design.

\begin{figure}[!ht]
\begin{AIbox}{Interior Design: Small Living Room (1/2)}
\scalebox{0.81}{
\begin{tabular}{p{1.1cm} p{13.6cm}  }

User & \textcolor{magenta}{(Generate New Image) [\texttt{Language Instruction} = "Living room with sofa and good lighting"][Generate]} \\
Assistant & \includegraphics[height=4.0cm]{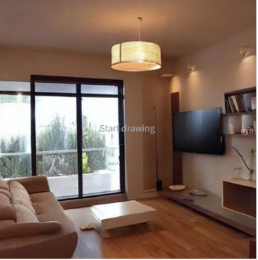} \\
User & How could I improve this room to make it more modern and cozy? \\
Assistant &
    To make the room more modern and cozy, you can consider the following improvements:
    \begin{enumerate}
        \item Upgrade the furniture: Replace the couch and chair with more contemporary designs, such as sleek lines and neutral colors. You can also add a stylish coffee table and side tables to complement the seating area.
        \item Update the lighting: Replace the current lighting fixtures with more modern and energy-efficient options, such as LED lights or smart lighting systems. You can also add floor lamps or table lamps to create a warm and inviting atmosphere.
        \item Enhance the window treatments: Replace the current window treatments with more modern and functional options, such as curtains, blinds, or shades. You can also consider adding a window seat or a bench to create a cozy seating area near the window.
        \item Add decorative elements: Introduce more decorative elements, such as artwork, plants, or decorative pillows, to make the room feel more personalized and welcoming.
        \item Incorporate technology: Add a smart TV or a sound system to the room to create a more modern and entertainment-focused space.
    \end{enumerate}
    By implementing these changes, you can transform the living room into a more modern and cozy space that reflects your personal style and preferences.  \\    

User & \textcolor{magenta}{(Remove or Change Object) [\texttt{Referral Text} = "sofa"] [Segment][\texttt{Grounding Instruction} ="Contemporary sofa"] [Generate]} \\
Assistant & \includegraphics[height=4.0cm]{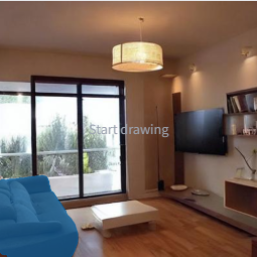} \includegraphics[height=4.0cm]{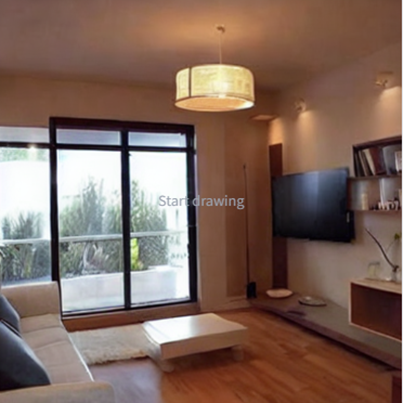}\\

User & \textcolor{magenta}{(Remove or Change Object) [\texttt{Referral Text} = "White table"] [Segment][\texttt{Grounding Instruction} = "modern coffee table"] [Generate]} \\
Assistant & \includegraphics[height=4.0cm]{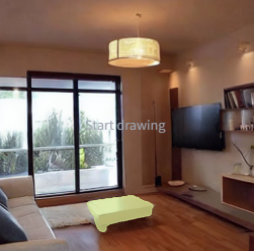} \includegraphics[height=4.0cm]{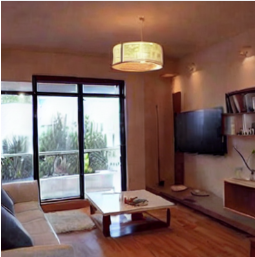}\\

User & \textcolor{magenta}{(Insert New Objects) [Draw box for \texttt{Grounding Instruction} = "Potted plant"] [Generate]} \\
Assistant & \includegraphics[height=4.0cm]{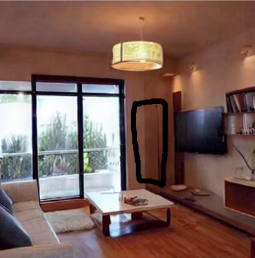} \includegraphics[height=4.0cm]{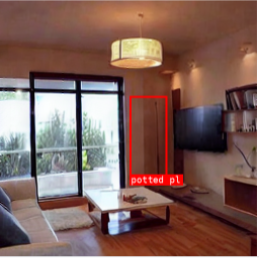} \includegraphics[height=4.0cm]{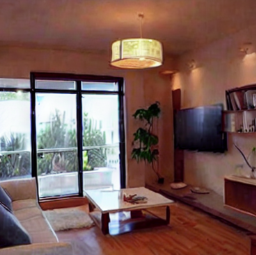}\\

\end{tabular}
}
\end{AIbox}
\vspace{-2mm}
\caption{
Interior Design: Small Living Room (1/2).  Refer to the scenario discussions in Section~\ref{sec:interior_design}.
} \label{fig:design_small_room_1}
\vspace{-0mm}
\end{figure}

\begin{figure}[!ht]
\begin{AIbox}{Interior Design: Small Living Room (2/2)}
\scalebox{0.83}{
\begin{tabular}{p{1.3cm} p{13.2cm}  }

User & I've changed some furniture in the room. Do you think it is good idea to lower the TV? \\
Assistant &
    Yes, lowering the TV in the living room is a good idea. In the image, the TV is currently mounted on the wall at a higher position. Lowering the TV to a more comfortable viewing height can improve the overall viewing experience for the people in the room. This can help reduce strain on the neck and eyes, especially when watching the TV from the couch or other seating areas. Additionally, lowering the TV can make it easier for people to see the screen, especially if they are sitting at a lower angle or distance from the TV. \\    

User & \textcolor{magenta}{(Remove or Change Object) [\texttt{Referral Text} = "TV"] [Segment][Drag segmented object to a lower position][\texttt{Grounding Instruction} = "TV"] [Generate]} \\
Assistant & \includegraphics[height=4.0cm]{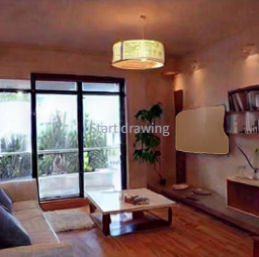} \includegraphics[height=4.0cm]{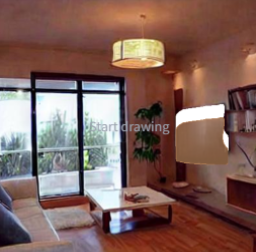} \includegraphics[height=4.0cm]{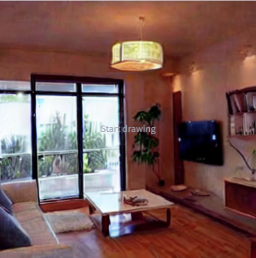}\\

User & I would like to replace the ceiling lamp. Do you think it is good idea? \\
Assistant & 
    Yes, it could be a good idea to replace the ceiling lamp in the living room. The current lamp appears to be old and has a yellowish hue, which might not provide the desired lighting effect or ambiance. Replacing it with a modern, energy-efficient lamp or a more visually appealing design can enhance the overall aesthetics of the living room and create a more comfortable and inviting atmosphere. Additionally, upgrading the lighting can also improve the room's functionality and energy efficiency.\\

User & \textcolor{magenta}{(Remove or Change Object) [\texttt{Referral Text} = "Lamp"] [Segment][\texttt{Grounding Instruction} = = "contemporary chandelier"] [Generate]} \\
Assistant & \includegraphics[height=6.0cm]{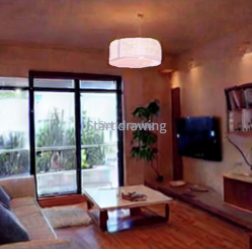} \includegraphics[height=6.0cm]{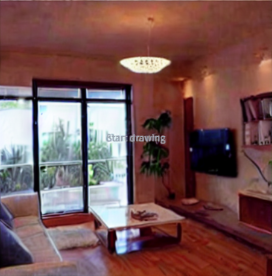}\\

\end{tabular}
}
\end{AIbox}
\vspace{-2mm}
\caption{
Interior Design: Small Living Room (2/2).  Refer to the scenario discussions in Section~\ref{sec:interior_design}.
} \label{fig:design_small_room_2}
\vspace{-0mm}
\end{figure}

\paragraph{Interior Design: Small Living Room.}
Please see the user-assistant interaction in 
Figure~\ref{fig:design_small_room_1} and ~\ref{fig:design_small_room_2}.
In a different living room design scenario, the assistant offers an extensive list of improvements. The user utilizes the interactive segmentation ability of the AI to select the sofa and table, updating them to a modern style. Additionally, a potted plant is incorporated using the grounded inpainting skill. The user might also inquire about specific objects, such as the position of the TV or replacing the ceiling lamp. Based on the suggestions, these objects can be modified using segmentation and inpainting skills to achieve the desired outcome.

\subsection{Identification of Unusual and Risky Items}
\label{sec:risky_items}
\begin{figure}[!ht]
\begin{AIbox}{Identification of Unusual and Risky Items.}
\scalebox{0.83}{
\begin{tabular}{p{1.3cm} p{13.2cm}  }

User & 
\includegraphics[height=5.15cm]{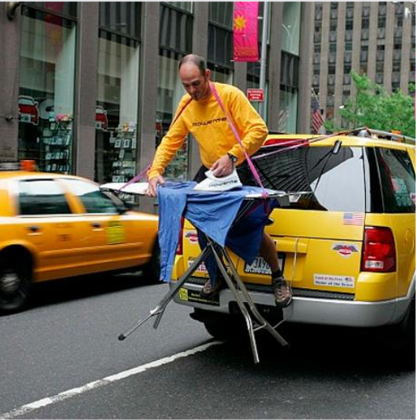} \\
User & what is unusual about this image? is it dangerous? \\
Assistant &
The unusual aspect of this image is that a man is ironing clothes while standing on the back of a moving car. This is dangerous because ironing clothes typically requires a stable surface and proper safety measures, such as using an ironing board and wearing appropriate protective gear. Standing on the back of a moving car is not only unstable but also poses a significant risk to the man's safety, as he could lose his balance and fall off the car, potentially causing severe injuries. Additionally, the man's actions could distract the driver, increasing the risk of an accident. \\    

User & \textcolor{magenta}{(Remove or Change Object) [\texttt{Referral Text} = "person"][Visual prompt by stroking on "leg"] [Segment][Drag segmented objects to remove] [Generate]} \\
Assistant & \includegraphics[height=5.15cm]{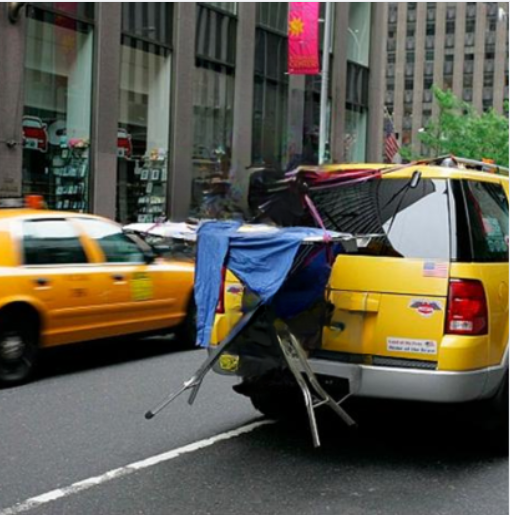}\\

User &  what is unusual about this image? is it dangerous? \\
Assistant & 
The unusual aspect of this image is that a yellow SUV is driving down the street with a chair strapped to its roof. This is not a typical sight, as it is not common for people to transport furniture, such as chairs, on the roof of their vehicles while driving. The situation could be potentially dangerous, as the chair might not be securely fastened, and it could fall off the vehicle while in motion, causing accidents or damage to the chair and other vehicles on the road. Additionally, the driver's visibility might be obstructed by the chair, making it difficult to see other vehicles or pedestrians, which could also lead to accidents. \\

User & \textcolor{magenta}{(Remove or Change Object) [Visual prompt by stroking on "yellow taxi"] [Segment][\texttt{Grounding Instruction} = "flowers"] [Generate]} \\
Assistant & \includegraphics[height=5.15cm]{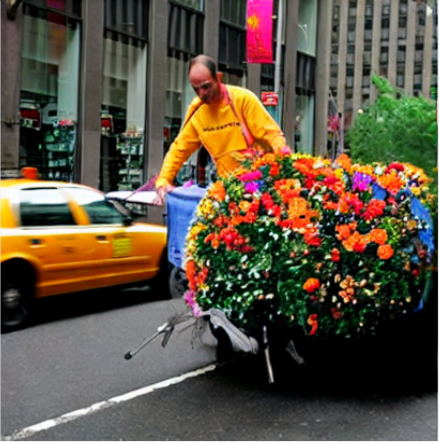}  \\

User & what is unusual about this image? is it dangerous? \\
Assistant & 
The unusual aspect of this image is the man riding a bicycle that is carrying a large, colorful floral arrangement, which appears to be a giant bouquet of flowers. This is not a common sight, as people typically do not transport such large and heavy objects on bicycles. The man is also wearing a yellow shirt, which adds to the uniqueness of the scene. The presence of a car in the background further emphasizes the contrast between the man's bicycle and the typical mode of transportation seen on the street. The combination of the man's attire, the bicycle, and the giant bouquet of flowers creates an interesting and eye-catching image. \\

\end{tabular}
}
\end{AIbox}
\vspace{-2mm}
\caption{
Identification of Unusual and Dangerous Items.  Refer to the scenario discussions in Section~\ref{sec:risky_items}.
} \label{fig:extreme_ironing}
\vspace{-0mm}
\end{figure}

Please see the user-assistant interaction in Figure~\ref{fig:extreme_ironing}. 
It is an important aspect of safety and security to detect the unusual, abnormal and risky items in images. The process typically involves analyzing images to detect any objects or patterns that may pose a risk or deviate from the norm. We use the popular extreme ironing image as an example to illustrate this use scenario. A typical question is to report what is unusual in the image. We further ask about the potential dangers. \shortname{} is able to correctly respond to this due to the use of the underlying LLaVA model. The image appears to be unusual because of the co-existence of the moving car and the man doing ironing. We ablate this by removing one element at each time. By removing the person, the ironing activity is not reported. Instead, the presentation of a chair on a moving car becomes the key unusual element. By replacing the moving taxi with flowers, the moving car is not reported. Instead, the assistant perceives the unusual aspect as that the man is riding a bicycle that is carrying a large, colorful bouquet of flowers. This is possible as the riding bicycle might be hiding behind the flowers. This analysis-by-synthesis approach can be effective in examining visual scenes to identify anomalies.

\section{Conclusions and Future Work}
In this paper, we have introduced \shortname{}, a research demo prototype that showcases the practical applications of large multimodal models featuring visual input, output, and interaction. \shortname{} is cost-effective in system development since it combines three pre-trained multimodal models of complementary skills using web services, without requiring additional model training: LLaVA for visual chat, SEEM for interactive image segmentation, and GLIGEN for grounded image generation and editing. At the system level, compared with other systems, \shortname{} is a fully vision-language multimodal system in terms of input, output, and interaction, particularly unique in supporting visual prompts for image segmentation and generation/editing. Our initial assessment of \shortname{} across a wide range of real-world application scenarios has demonstrated its excellent ability to perform new, complex tasks. We hope this will inspire further research into multimodal foundation models.

We identify several potential avenues for future research:
$(i)$ The abilities of \shortname{} are restricted by the performance limits of the utilized pre-trained models. Enhancing \shortname{}'s specific skill could be achieved by replacing the module with a superior model variant or creating an improved individual model, such as LLaVA, SEEM, and GLIGEN. System development and individual model development can be de-coupled, allowing for a plug-and-play approach to system serving. We also hope to extend the system development framework by incorporating additional features like Instruct Pix2Pix~\citep{brooks2023instructpix2pix} for image-level editing.
$(ii)$ Since \shortname{} is a composite of individual models, its capacity during each inference is determined by the existing abilities of those models. While more complex tasks can be accomplished through iterative activation of current skills for combined skills, no new skill emerges at each inference by interpolating in the neural network's hidden space. We encourage the community to develop multimodal foundation models with more unified modeling, allowing new capabilities to emerge through latent task composition.

\begin{ack}
The authors would like to express their gratitude to the MSR Central Engineering Team for their support. We appreciate the efforts of all authors who contributed to the individual pre-trained models, making \shortname{} a reality. Special thanks go to Matt Mazzola for creating the informative demo instruction video and managing the demo serving, as well as Swadheen Shukla and Lars Liden for their valuable input and insightful discussions.
\end{ack}

\bibliography{egbib}
\bibliographystyle{plain}

\begin{CJK*}{UTF8}{gbsn}
\end{CJK*}

\begin{CJK*}{UTF8}{gbsn}
\end{CJK*}

\begin{CJK*}{UTF8}{gbsn}
\end{CJK*}

\begin{CJK*}{UTF8}{gbsn}
\end{CJK*}

\begin{CJK*}{UTF8}{gbsn}

\end{CJK*}

\begin{CJK*}{UTF8}{gbsn}

\end{CJK*}

\begin{figure}[!ht]
\begin{AIbox}{Recipe (2/2)}
\scalebox{0.83}{
\begin{tabular}{p{1.3cm} p{13.2cm}  }

    User &  I updated the picture adding rice because it is my favorite food. Could you please give me a new recipe including rice. \\
    Assistant & Sure, here's a new recipe that includes rice:\newline
    Ingredients:
    \begin{enumerate}    
        \item 1 lb. beef (sliced into thin strips)
        \item 4 carrots (sliced)
        \item 2 potatoes (peeled and cubed)
        \item 1 cup white rice
        \item 1/4 cup chopped onion
        \item 2 cloves garlic, minced
        \item 1/4 cup soy sauce
        \item 2 tbsp. olive oil
        \item 1 tsp. sesame oil
        \item 1 tsp. cornstarch
        \item Salt and pepper to taste
        \item Chopped green onions for garnish
        \end{enumerate}
    Instructions:
    \begin{enumerate}
        \item In a small bowl, whisk together the soy sauce, olive oil, sesame oil, and cornstarch. Set aside.
        \item Heat a large skillet or wok over medium-high heat. Add the beef and cook until browned, about 5-7 minutes. Remove the beef from the skillet and set aside.
        \item In the same skillet, add the onion and garlic and cook until the onion is translucent, about 2-3 minutes.
        \item Add the carrots and potatoes to the skillet and cook for another 5-7 minutes, or until the vegetables are tender but still crisp.
        \item Return the beef to the skillet and add the soy sauce mixture. Stir to combine and cook for another 2-3 minutes, or until the sauce has thickened.
        \item Add the rice to the skillet and stir to combine with the beef and vegetables. Cook for another 2-3 minutes, or until the rice is heated through.
        \item Season with salt and pepper to taste.
        \item Serve the beef and rice mixture over a bed of rice. Garnish with chopped green onions.        
    \end{enumerate}
    Enjoy your delicious and hearty rice dish!
    
\end{tabular}
}
\end{AIbox}
\vspace{-2mm}
\caption{
 Recipe (2/2).  Refer to the scenario discussions in Section~\ref{sec:dinner_preparation}.
} \label{fig:dinner_preparation2_2}
\vspace{-0mm}
\end{figure}

\begin{CJK*}{UTF8}{gbsn}
\end{CJK*}

\begin{CJK*}{UTF8}{gbsn}
\end{CJK*}

\begin{CJK*}{UTF8}{gbsn}
\end{CJK*}

\begin{CJK*}{UTF8}{gbsn}
\end{CJK*}

\begin{CJK*}{UTF8}{gbsn}
\end{CJK*}

\begin{CJK*}{UTF8}{gbsn}
\end{CJK*}

\begin{CJK*}{UTF8}{gbsn}
\end{CJK*}

\begin{CJK*}{UTF8}{gbsn}
\end{CJK*}


\end{document}